\documentclass{article} 
\usepackage{iclr2025_conference,times}

\usepackage{hyperref}
\usepackage{url}

\usepackage{graphicx}
\usepackage{algorithm}
\usepackage{algorithmic}
\usepackage{subcaption}
\usepackage{makecell}
\usepackage{tabularx}

\newtheorem{theorem}{Theorem}
\newtheorem{definition}{Definition}

\title{A Proposal for Networks Capable of \\ Continual Learning}

\author{Zeki Doruk Erden \& Boi Faltings\\
Artificial Intelligence Laboratory\\
École Polytechnique Fédérale de Lausanne\\
\texttt{\{zeki.erden, boi.faltings\}@epfl.ch}\\
}

\iclrfinalcopy 
\begin{document}

\maketitle

\begin{abstract}
    We analyze the ability of computational units to retain past responses after parameter updates, a key property for system-wide continual learning. Neural networks trained with gradient descent lack this capability, prompting us to propose \textit{Modelleyen}, an alternative approach with inherent response preservation. We demonstrate through experiments on modeling the dynamics of a simple environment and on MNIST that, despite increased computational complexity and some representational limitations at its current stage, Modelleyen achieves continual learning without relying on sample replay or predefined task boundaries.
\end{abstract}

\section{Introduction}
\label{sec:introduction}

Modern machine learning relies on neural networks (NNs) to model complex systems, achieving remarkable success in areas like image recognition, language processing, and complex behavior. Yet as these problems are solved, crucial shortcomings at their core start to gain attention (\cite{clune2019ai,zador2019critique,marcus2018deep,lecun2022path}). Importantly, they struggle with continual learning—the ability to learn new tasks without forgetting previous ones, as essential for open-ended learning, reflecting real-world dynamics. This limitation necessitates frequent, resource-intensive retraining, making scalability impractical. In this paper, we analyze a property essential for system-wide continual learning: the ability of computational units to retain past responses after parameter updates. We show that neurons in gradient-based NNs lack this capability and propose the early design of an alternative approach, \textit{Modelleyen}, where units inherently preserve past responses. We also extend this approach to larger networks and, through experiments on modeling a simple environment and on MNIST, demonstrate that despite higher computational complexity and some performance limitations due to representational limitations at its current stage, we achieve system-wide continual learning without requiring replay or task boundaries. It also offers the advantage of human-comprehensible internal representations, as shown in additional results.

\section{Related work}
\label{sec:related_work}

The inability of continual learning is a core limitation of current ML systems  (\cite{hadsell2020embracing,zhuang2020comprehensive,qu2021recent}). Solutions proposed for for this problem often don't fully resolve the fundamental limitations of NNs in this regard but aim to mitigate their effects. Many continual learning methods rely on simplifying assumptions, like externally defined task boundaries and task change information (\cite{rusu2016progressive, jacobson2022task, wang2022continual}, or storing and replaying past observations \cite{buzzega2020dark, galashov2023continually}), which bias learning toward previous tasks without enabling true continual learning (\cite{kirkpatrick2017overcoming}). One exception is methods generating multiple experts and assigning tasks based on a generative model (\cite{erden2024directed, lee2020neural}), though these still assume distinct tasks by decomposing the system into experts, relying on the availability of bulk data for each expert’s generative model – an assumption that does not hold in real-world learning. Thus, the continual learning problem merely shifts to the generative model for task identification rather than the entire system.

Over the past decades, evolutionary biology has highlighted the crucial role of local variation and selection in shaping biological structures and driving their evolution \cite{gerhart2007theory,marc2005plausibility,west2003developmental}, including neural processes in the brain \cite{marc2005plausibility,edelman1993neural}. Our method centers on this mechanism, which is vital for achieving continual learning.

\section{Local preservation of past responses}

System-wide preservation of past response is difficult to define and analyse, likely impossible without explicit retained knowledge of past samples. For that, we turn our attention to a local preservation of past response, which acts as a useful tool for system-wide continual learning, if not a prerequisite.

We look at computational units whose response $y$ can be phrased as $y=f(\sum w_i x_i + b)$ where $w_i$ are the weights, $x_i$ are the inputs and $b$ is the bias of the unit, and $f(\cdot)$ is a nonlinear activation function. Neurons in NNs are examples of such units (as well as CSVs in our design, see next section).

Assume that the unit, in the past, has been exposed to an input instance $\textbf{X}_t = [x_{0t}, ... x_{Nt}]$ at some step $t$. With the standing weights $\textbf{W} = [w_0, w_1, ... w_N]$, its response to this input given as $y_t=f(\sum w_i \cdot x_{it} + b)$. Suppose that after an arbitrary process of learning in response to other input instances, the parameters of the unit are updated to $\textbf{W}' = [w'_0, ... w'N]$ $w'_i = w_i + \Delta w_i$ and $b' = b_i + \Delta b$ at step $t=T$. New response to past input $\textbf{X}_t$ is now $y'_t = f(\sum w'_i x_{it} + b')$. We want to ensure $y'_t=y$, or  $f(\sum w'_i x_{it} + b') = f(\sum w_i x_{it} + b)$. This is generally valid (and valid for monotonous activations used in NNs only) if $ \sum (w_i + \Delta w_i) x_{it} + (b + \Delta b) = \sum w_i x_{it} + b$, or if:

\begin{equation}
    \label{eq:popr_condition}
    \sum \Delta w_i x_{it} + \Delta b = 0
\end{equation}

Eq. \ref{eq:popr_condition} is the condition for local preservation of information. A learning update rule defining $\Delta w_i$ and $\Delta b$ should satisfy this condition if response of the unit to previous inputs $\textbf{X}_t$, $t \leq T$ is to be preserved. 

Some of the terms in the sum of Eq. \ref{eq:popr_condition} may be $0$ due to $\Delta w_i$ being 0, as defined by the learning rule, based on the current value of $w_i$. So we can rewrite \ref{eq:popr_condition} as:

\begin{equation}
    \sum_{i:\Delta w_i \neq 0} \Delta w_i x_{it} + \Delta b = 0
\end{equation}

Note that this equation is in general dependent on $x_{it}$ of $t \leq T$ for $i:\Delta w_i \neq 0$, which we assume to be not explicitly available. The only possibility for an update rule to satisfy the condition is to make it independent of $x_{it}$. This would only be possible if $\forall i:\Delta w_i \neq 0$, and $\forall t \leq T$, $x_{it}$ can be readily deduced from the available information, in which case $w_i$. This can be written as:

\begin{equation}
    \label{eq:x_g_rel}
     \forall i:\Delta w_i \neq 0,\ \forall t \leq T,\  w_i = g(x_{it})\rightarrow x_{it} = g^{-1}(w_i)
\end{equation}

So that \ref{eq:popr_condition} becomes independent of $x_{it}$: $\sum_{i:\Delta w_i \neq 0} \Delta w_i g^{-1}(w_i) + \Delta b = 0$. It follows from \ref{eq:x_g_rel} that:
   
\begin{equation}
    \label{eq:xs_rel}
    x_{i0} = x_{i1} = ... = x_{iT},\ \forall t \leq T,  \forall i:\Delta w_i \neq 0\
\end{equation}

These conditions do not apply to neural networks trained by gradient descent because there is (1) no direct correspondence between a learned weight and the input value it modulates (Eq. \ref{eq:x_g_rel}), and (2) no guarantee that previously encountered inputs will be identical for a given weight (Eq. \ref{eq:xs_rel}). In contrast, our design meets these conditions, as outlined at the end of the following section.

\section{Modelleyen: Learning basic environment dynamics}

Our proposal, termed \textit{Modelleyen}, is designed to model sequential observations from an environment, but can be applied to any prediction task. It learns the environment’s structure with minimal exposure, enabling information reuse and continual learning while maintaining consistency with past experiences. At the core of our method is a local variation and selection process \cite{gerhart2007theory} as essential for continual learning and structured environment modelling.

Below, we outline Modelleyen's core mechanism, which relies on the immediate succession of activities in discrete state variables to model simple environmental dynamics. The next section extends this mechanism to observation spaces that can be represented as networks. Due to space constraints, we provide an overview of key definitions, the basic learning process, and core continual learning properties. For a full description, see Appendix \ref{sec:modelleyen_details} and Algorithms \ref{alg:algorithm_adaptationloop} and \ref{alg:algorithm_csvstate}.

\begin{definition}
    (State Variable - SV) A state variable $X$ is a unit in our system whose state, $S_X$, can take values 1 (\textit{active}), -1 (\textit{inactive}), or 0 (\textit{unobserved/undefined} depending on context). 
\end{definition}

SVs can be interpreted as boolean variables with possibility to take an \textit{unobserved} value. The integers assigned for states are only for notation and not for algebraic operation. The following are subtypes of SVs:

\begin{definition}
    (Base SV - BSV and Dynamics SV - DSV) A \textit{BSV} $X$ is an SV whose values are provided externally each timestep and whose state is limited by $S_X\in\{-1,1\}$. Each BSV comes with two \textit{DSVs}, $X_A$ and $X_D$, that represent its activation and deactivation at current step ($t$) compared to previous timestep respectively; where $S_{X_A}=1$ if and only if $S_X(t-1)=-1 \land S_X(t)=1$, and $S_{X_D}=1$ if and only if $S_X(t-1)=1 \land S_X(t)=-1$, and persisting as long as no new event in BSVs are observed.
\end{definition}

\begin{definition}
    (Conditioning SV - CSV) A CSV $C$ is a type of SV with mutable sets of positive sources $X_P$, negative sources $X_N$, and conditioning targets $Y$. Positive and negative sources are BSVs and DSVs, while targets can be DSVs or other CSVs. The sources of $C$ are considered "satisfied" if all positive sources are active and all negative sources are not active. If sources are satisfied, $S_C=1$ if sources are satisfied and $S_Y \in \{0,1\},\ \forall x \in Y$ (targets are active); $S_C=-1$ if sources are satisfied and $S_Y \in \{0,-1\},\ \forall x \in Y$ (targets are inactive), and $S_C=0$ otherwise. Additionally, each CSV has a "unconditionality" flag, which indicates if the CSV has, in the past, been always observed active when sources were satisfied ("unconditional"), was never observed active without a predictive explanation ("conditional"), or was sometimes observed active without a predictive explanation ("possibly conditional"), the latter representing uncertainty in a qualitative manner.
    
\end{definition}

BSVs are essentially environment observations, while DSVs represent their changes.\footnote{In our implementation we also use BSVs to represent actions taken by the agent in the previous step. The actions do not have associated DSVs, since their activation and deactivation is in agent's control.} CSVs model the presence or absence of a relationship between a learned condition (sources) and its effect (active target states), indicated by the CSV being active (1) or inactive (-1). Figure \ref{fig:svs} in Appendix summarizes these SV types and their connections. Note that CSVs are \textit{not} feedforward computational units; they represent the relationship between sources and targets - states of their targets are set independently of the CSV, unlike feedforward units that determine target states based on source states. CSVs partially function as feedforward units only when used for prediction of alternative outcomes.

\begin{figure*}
     \centering
     \begin{subfigure}[t]{0.19\textwidth}
         \centering
         \includegraphics[width=\textwidth]{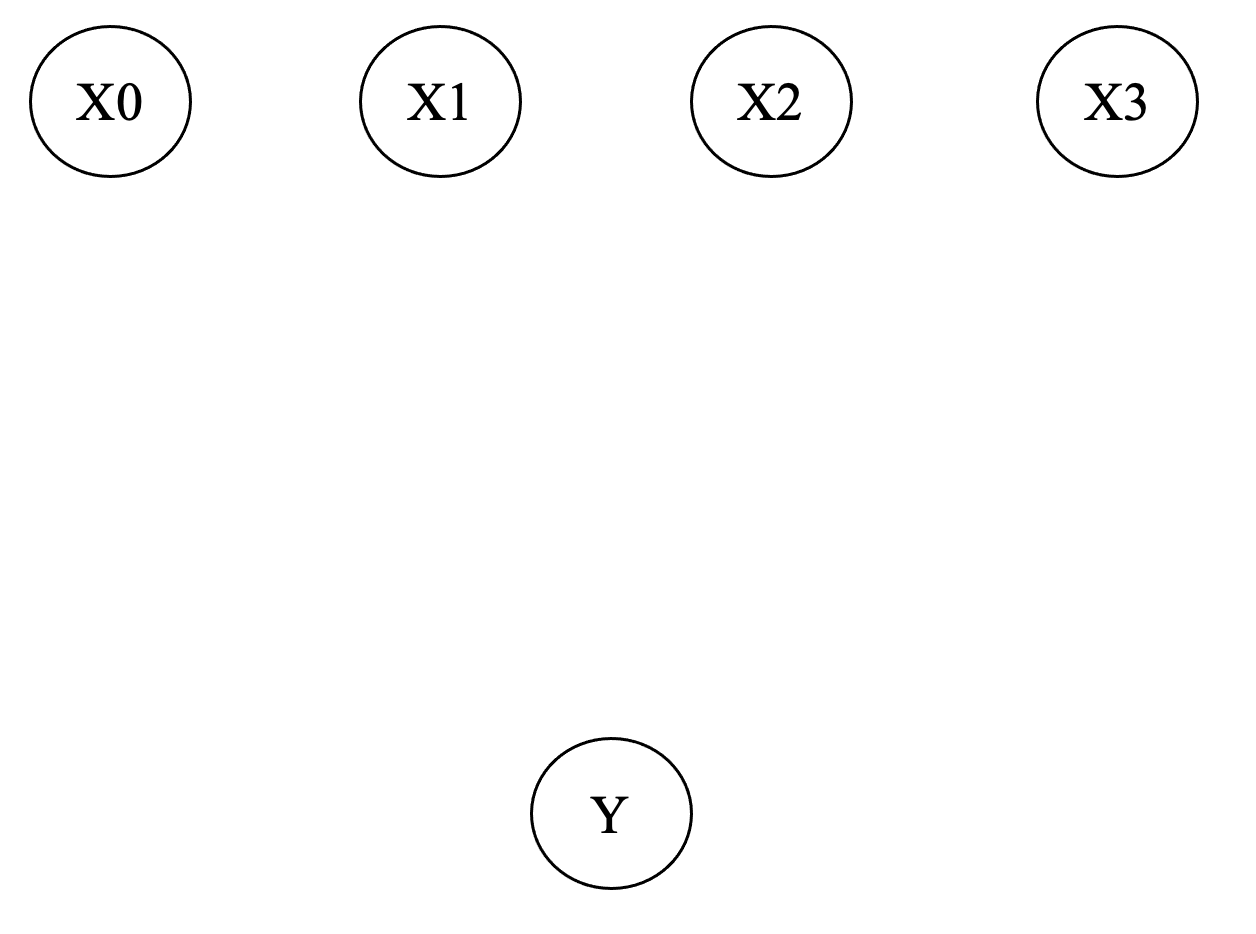}
         \caption{}
         \label{fig:csvform_1}
     \end{subfigure}
     \hfill
     \begin{subfigure}[t]{0.19\textwidth}
         \centering
         \includegraphics[width=\textwidth]{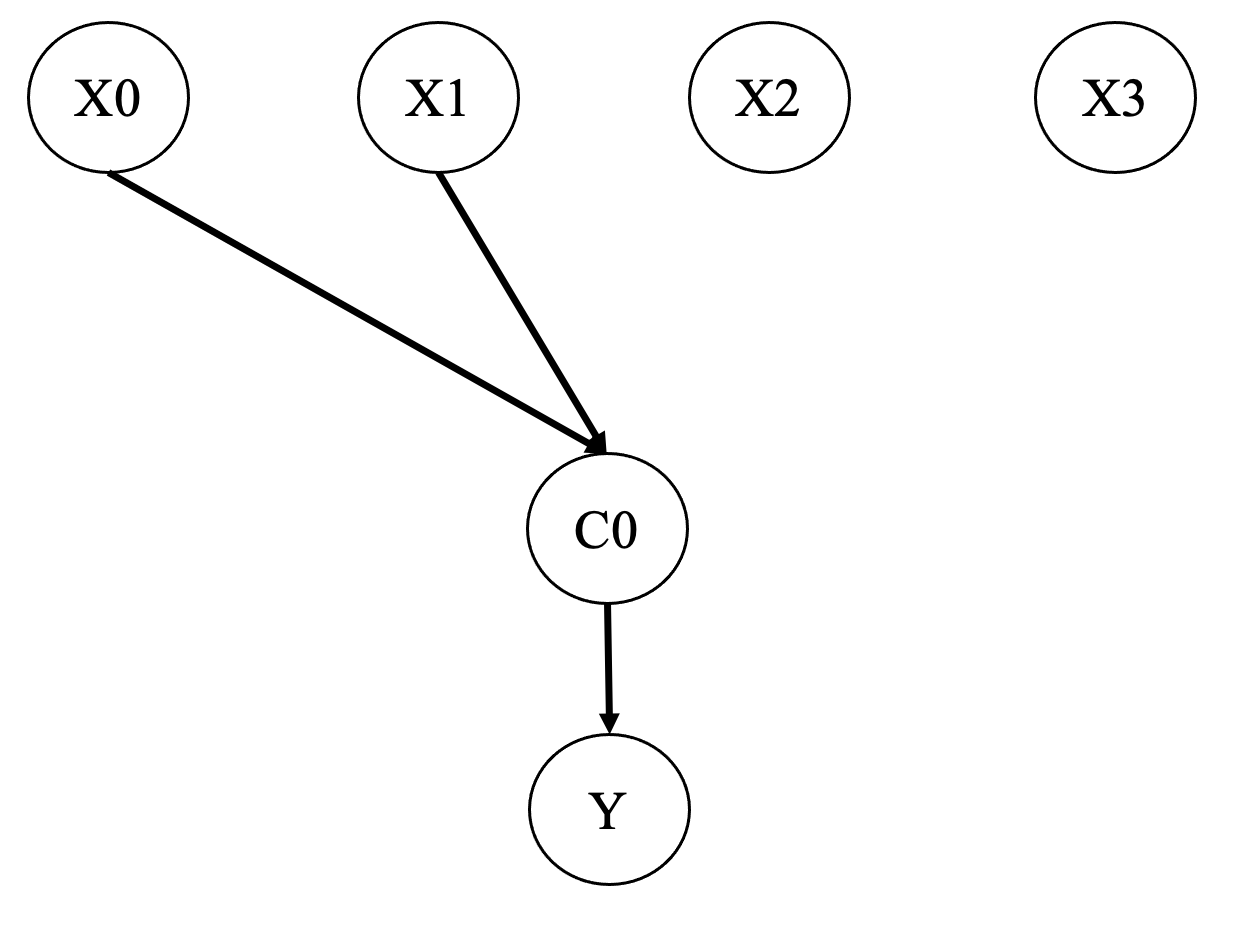}
         \caption{}
         \label{fig:csvform_2}
     \end{subfigure}
     \hfill
     \begin{subfigure}[t]{0.19\textwidth}
         \centering
         \includegraphics[width=\textwidth]{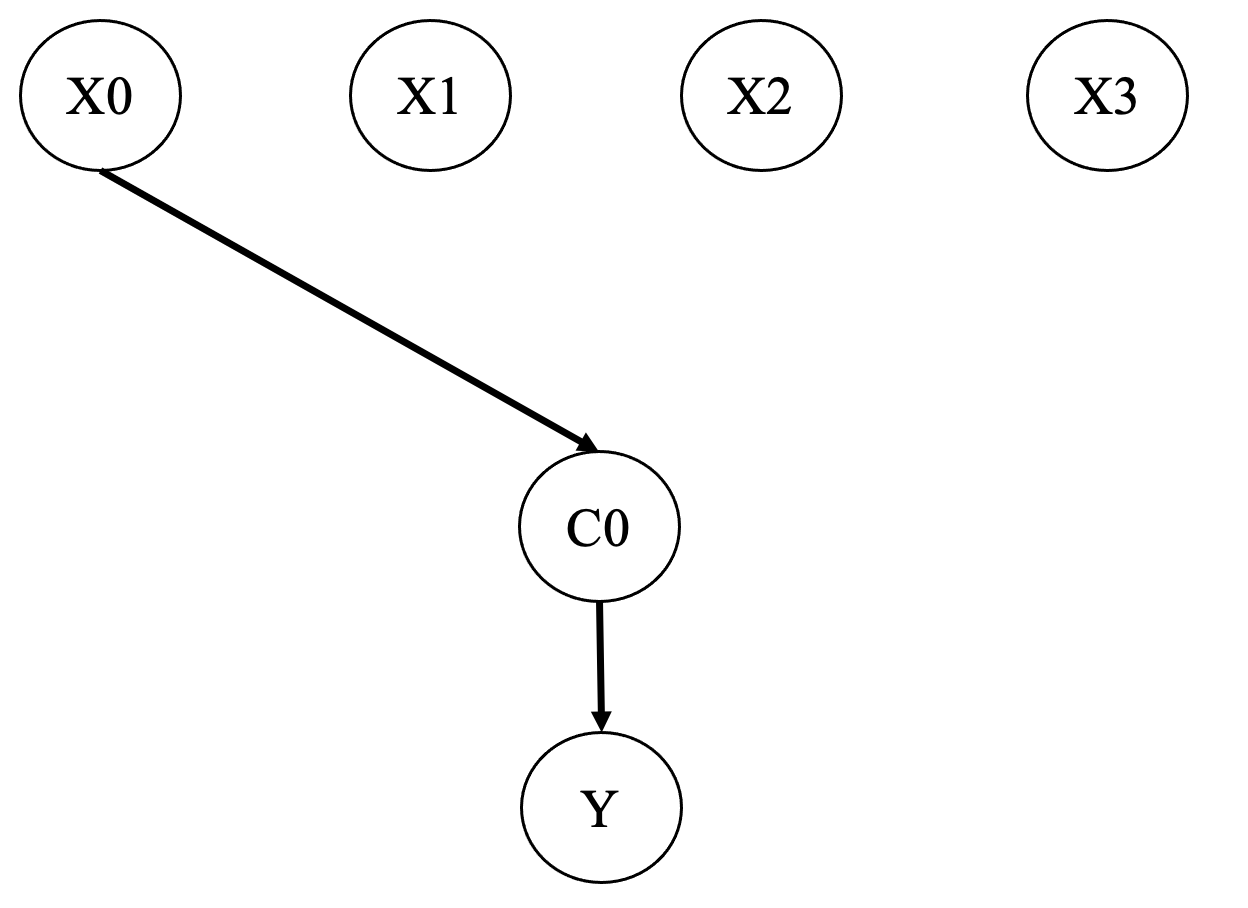}
         \caption{}
         \label{fig:csvform_3}
     \end{subfigure}
     \hfill
     \begin{subfigure}[t]{0.19\textwidth}
         \centering
         \includegraphics[width=\textwidth]{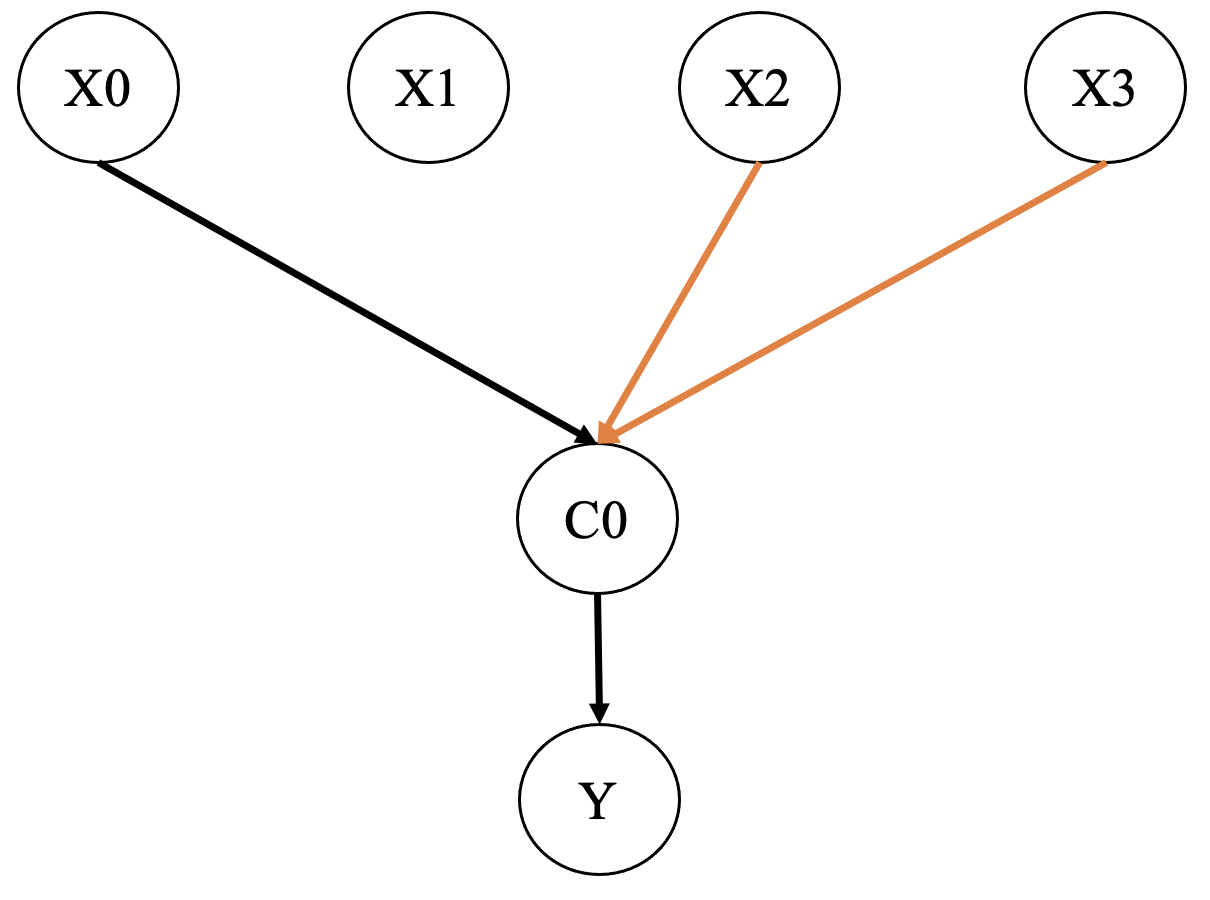}
         \caption{}
         \label{fig:csvform_4}
     \end{subfigure}
     \hfill
     \begin{subfigure}[t]{0.19\textwidth}
         \centering
         \includegraphics[width=\textwidth]{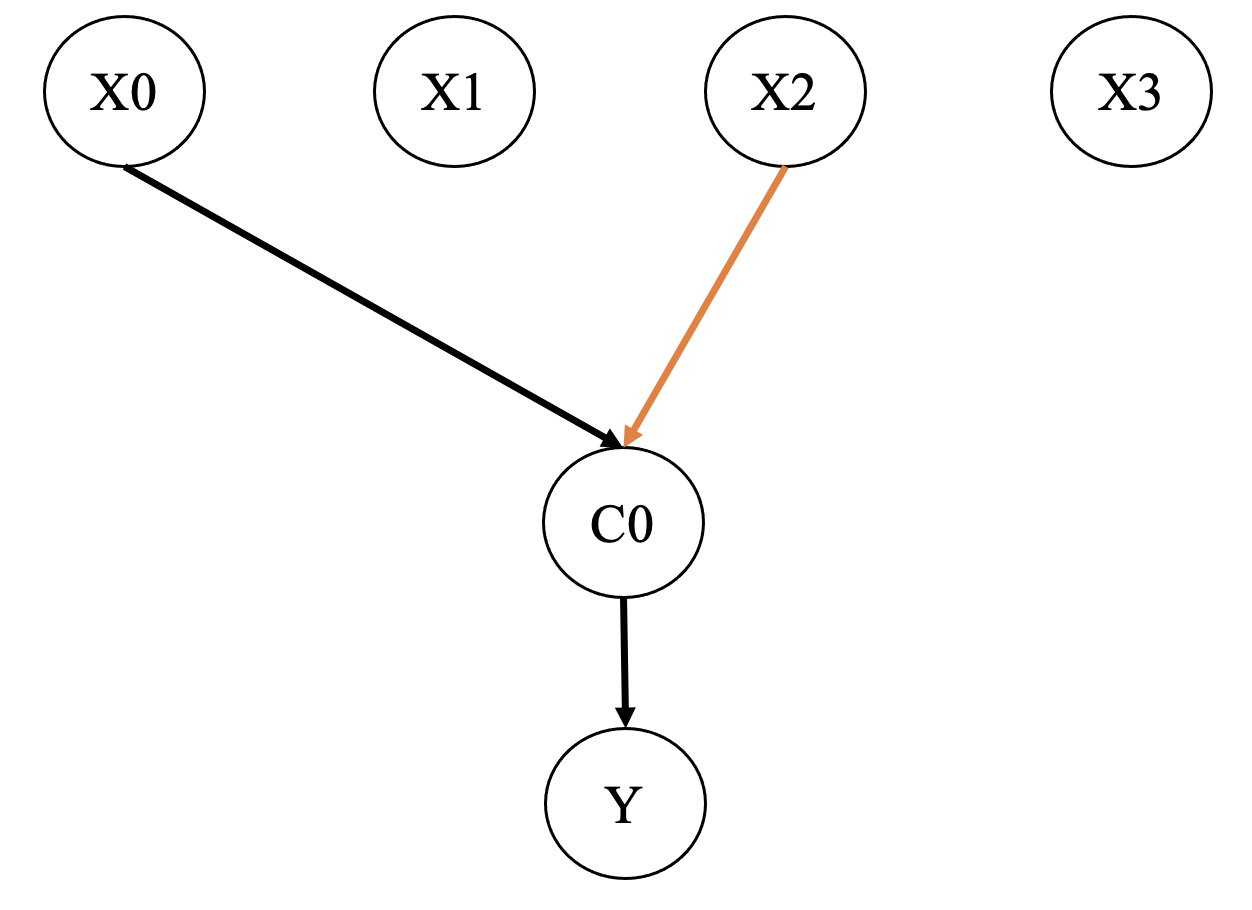}
         \caption{}
         \label{fig:csvform_5}
     \end{subfigure}
    \caption{Sample formation of a CSV in a continual manner. The relationship to be modelled is $Y = X0\ and\ !X2$ ("!" denotes "not"). Black and orange arrows represent positive and negative sources for CSV $C0$ respctively. $Xi$ can be interpreted either as single or grouped SVs. (a) Initial state with no relation formed between $X0-3$ and $Y$. (b) $X0, X1 \rightarrow Y$ observed. Positive connections hypothesizing both $X0$ \& $X1$ are required for Y are formed. (c) $X0 \rightarrow Y$ is observed. $X1$ is deduced unnecessary for $Y$. (d) $X0, X2, X3 \rightarrow !Y$ observed. $Y$ is hypothesized to be suppressed by $X2$ and $X3$. (e) $X0, X2 \rightarrow !Y$ observed. $X3$, seen unnecessary for suppression of $Y$, refined. Correct structure learned and is stable.}
    \label{fig:csvform}
\end{figure*}

The learning process proceeds \textit{step-by-step}, without incorporating an aggregate evaluation of multiple observational samples gathered from the environment, nor relying on repeated passes over a batch of data—distinct from traditional approaches. Initially, the model includes only BSVs and their DSVs, with no CSVs. At each step, Modelleyen seeks to explain the observed states of CSVs and DSVs in the previous timestep (modeling BSVs indirectly via DSVs). It does so by creating new CSVs to account for unexplained DSVs and CSVs. These retrospective explanations captured by CSVs become predictions for potential outcomes in the next timestep. Learning capability of Modelleyen comes from the \textit{operations} on CSVs - their formation, and the modification of their positive and negative sources; summarized as follows (detailed on Algorithms \ref{alg:algorithm_adaptationloop} and \ref{alg:algorithm_csvstate}):

\textit{Initial formation:} Figure \ref{fig:csvform_2}. At each step, if there are active DSVs or CSVs without an explanation (an active conditioner or an unconditionality flag, see Appendix), a new CSV is generated to explain them. Initially, the CSV has no negative sources ($X_N = \emptyset$) and includes all active BSVs and DSVs at that timestep as positive sources ($X_P$). No additional positive sources can be added to the CSV.

\textit{Negative connections formation:} Figure \ref{fig:csvform_4}. At the first instance where a CSV's sources are satisfied but its state is inactive, the CSV receives all active DSVs and BSVs at that timestep as negative sources ($X_N$), similar to previous step. No additional negative sources are added thereafter.\footnote{This process is separate from initial sources' formation to only to avoid exhaustive negative connections.}

\textit{Refinements:} Figures \ref{fig:csvform_3} and \ref{fig:csvform_5}. When a CSV's state is determined as 1 with at least one active positive source and active targets, we remove nonactive positive sources (${x \in X_P : S_X \neq 1}$) from $X_P$ and active negative sources (${x \in X_N : S_X = 1}$) from $X_N$. When the state is 0, with at least one active positive source, inactive targets, and at least one active negative source, we remove nonactive negative sources (${x \in X_N : S_X \neq 1}$) from $X_N$.

Intuitively, a CSV starts by being connected to all active SVs at formation, representing a comprehensive hypothesis of relationships. These relationships are then refined based on observations where some connections are deemed unnecessary, ensuring the CSV remains consistent with past observations locally. This refinement is central to Modelleyen's continual learning ability, evident from its lowest organizational level of CSVs, as formalized of the following property.

\begin{theorem}
    Let $y_i$ be an \textit{instance} that includes the previous states of all the positive and negative sources of a CSV $C$ and the current states of all its conditioning targets. Then, if $C$ undergoes any modification as a result of encounter with an instance $y_1$, its state in reponse to any past instance $y_0$ is not altered by this modification; as long as its targets remain identical and $C$ does not undergo negative sources formation.\footnote{The requirement for identicality of targets in this theorem is only to account for the fact that heterogeneous targets result in duplication of CSVs - see the Appendix for details of this mechanism. The theorem holds when one considers the response of the duplicated CSVs with respect to the targets assigned to each duplicate as well.} For the proof, see Appendix \ref{sec:app_proof}.
\end{theorem}

Theorem 1 is exemplified in Figure \ref{fig:csvform}: In \ref{fig:csvform_2}, after elimination of $X1$ as a positive source, the earlier exposure of $X0, X1 \rightarrow Y$ still results in a state of activity in $C0$, and likewise for $X2$ \& $X3$. With this property, we know that the state of a CSV in response to any past encounter is not altered except possibly for initial negative sources formation (happening only once per CSV), hence realizing continual learning without destructive adaptation in Modelleyen inherently and from the lowest level of organization.

Another way to look at Theorem 1 is it being a satisfaction of the conditions on \ref{eq:x_g_rel} and \ref{eq:xs_rel} by the following equivalences by CSVs as computational units: (1) Sources to which the CSV is no longer connected to are the instances that took different values in the past (equivalent to having $w_i=0$), and they are known to have $\Delta w = 0$ since a removed source remains disconnected. (2) For all the remaining connections ($\Delta w$ not necessarily $= 0$, i.e. can undergo change) we know that $x_{it}=1,\ \forall t \leq T$ since the ongoing presence of an input $i$ means that it has not been observed absent in the past. (3) The observation condition of a CSV (same as the condition for the satisfaction of its sources) is equivalent to $f(.)$ being a step function, $w_i$ being 1 for all connections, and $b = -|\textbf{W}_c|$, $\textbf{W}_c$ being the set of all connected sources. Hence, Modelleyen locally preserves past responses.

\begin{figure}
     \centering
     \begin{subfigure}[t]{0.12\textwidth}
         \centering
         \includegraphics[width=\textwidth]{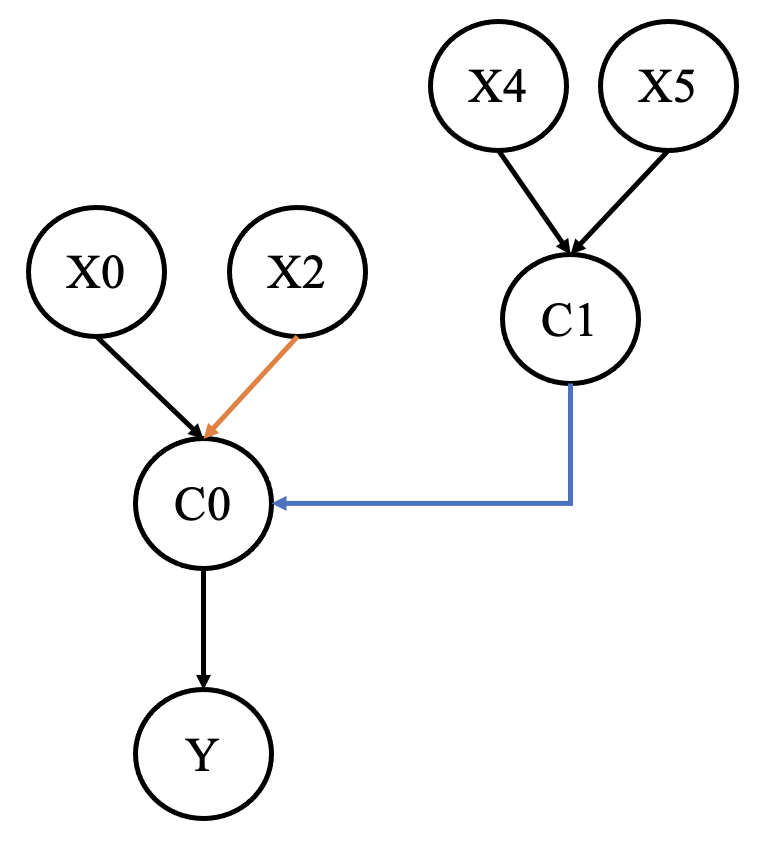}
         \caption{}
         \label{fig:csvformupstream_1}
     \end{subfigure}
     \vline
     \begin{subfigure}[t]{0.24\textwidth}
         \centering
         \includegraphics[width=\textwidth]{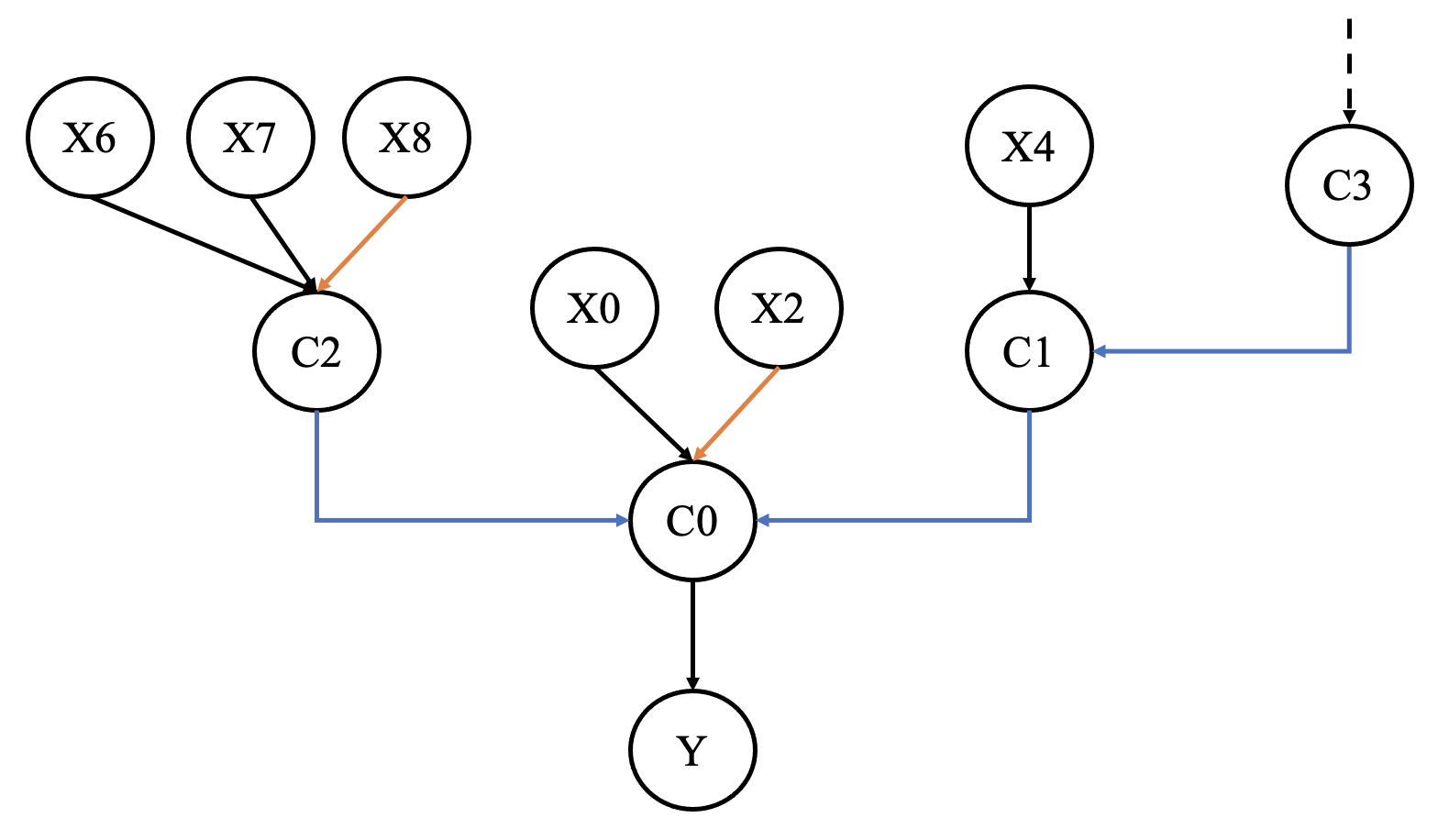}
         \caption{}
         \label{fig:csvformupstream_2}
     \end{subfigure}
    \caption{Example of upstream conditioning. In Figure \ref{fig:csvform}, assume the unconditionality flag (see Appendix \ref{sec:modelleyen_details}) of $C0$ is set after observing that $(X0, !X2)$ did not activate it. (a) Upon observing $X0, !X2, X4, X5 \rightarrow Y$, $C0$ is active, as $X0, !X2$ led to $Y$. A new CSV $C1$ forms and conditions $C0$. Note that $(X4, X5)$ alone won't activate $C0$ unless its sources are active. (b) New conditioners undergo CSV processes: $X5$ from $C1$ is refined, forming $C2$ and $C3$. Multiple conditioners represent alternatives, so $C0$ is activated when either $C1$ or $C2$ sources are active. This allows logical functions to be incorporated in a minimal, ongoing manner without losing past knowledge.}
    \label{fig:csvformupstream}
\end{figure}

A CSV can condition/predict not only the activation of direct environmental dynamics (DSVs), but also possibly the activation of other CSVs. This capability enables the model to become more complex upstream, allowing for the representation of arbitrarily complex logical relations in a structurally minimal way, without requiring any \textit{a priori} knowledge of the existence of such relations. (As a result, it is not constrained by the assumptions such as those in \cite{mordoch2023learning} discussed in the introduction). This formation of upstream conditioning pathways is exemplified on Figure \ref{fig:csvformupstream}, continuing our example from Figure \ref{fig:csvform}. The processes of refinements, negative sources formations, and even further upstream conditioning are identical regardless of what the target of a CSV is.

Additionally, in Appendix \ref{sec:statistical_significance}, we quantify the statistical significance of relationships between each CSV and their targets - this prevents excessive complexification in environments with numerous observations and spurious relationships, expected to be especially important when scaling to higher-dimensional environments. Details of this mechanism are excluded from the main text for brevity.

This approach fundamentally differs from methods like NNs. In Modelleyen, the agent updates its model instantly with each new observation, initially "overfitting" to fully account for data before refining it to be as simple and explanatory as possible without contradicting prior experience. The model remains as general as needed based on past exposures, but no more. A more specific representation (more sources per CSV) enables precise generalization with new observations, increasing consistency as sources are refined. This process mirrors biological systems, where redundancy allows for adaptive selection \cite{gerhart2007theory}. We propose calling such mechanisms —which rely on local variation and selection— \textit{varsel mechanisms}, and networks using them \textit{varsel networks}. Unlike conventional methods that avoid overfitting through gradual adjustment, varsel networks inherently build the necessary generalization from all prior data.

\section{Modelleyen with network refinement}

Our initial experiments with the base Modelleyen mechanism (Section \ref{sec:results}) use a low-dimensional finite state machine. While Modelleyen can theoretically handle any observation space, large spaces lead to overly complex models. Here, we provide an extension, called \textit{Modelleyen with Network Refinement (MNR)}, enabling it to operate on observation spaces represented as networks. While MNR generalizes to any network-representable data with proper conversion, our attention here is on vision, and 2D shape identification in particular. Details of our feature representation (how we convert the images to a network), one of many conceivable alternatives, are in Section \ref{sec:feature_representation}; though the algorithm is agnostic to the specific feature representation used and here, the reader can think about a generic concept of a visual feature that can be edges, gradients, objects, or raw pixels.

First, we define the basis of our representation of observations \& sources for CSVs in this extension:

\begin{definition}
    A \textbf{state network (SN)} is a directed graph $(N,E)$, where each node has a \textbf{type}. A list of state networks and associated keys, $P=[(k_0, SN_0), (k_1, SN_1), ...]$ is a \textbf{state polynetwork (SPN).}
\end{definition}

Node types in state networks 
(SNs) represent distinct features (e.g., edges, corners, or objects in visual space), with nodes being observed instances of these features in the current observation (e.g. edges with the same orientation or instances of the same object are distinct nodes of the same type; see Fig. \ref{fig:assignment_source} in Appendix for an example). Edges ($E$) represent relations between nodes, e.g. relative positions in visual inputs or succession in temporal domains. A state polynetwork (SPN) is a collection of distinct state networks with a designator key, enabling the definition of different feature and relation types. In visual space, this could include shape, color gradients, or abstract objects, as well as multi-dimensional relations (e.g., relative positioning). An example SPN is shown in Fig. \ref{fig:representation_example_vn}.

SPNs will serve as input sources for CSVs in our model, replacing the sets of state variables in base Modelleyen. Learning the model involves constructing SPN structures that capture the desired information, bringing us to the question of how an SPN (representing a CSV's input configuration) should be modified in response to new SPN observations.

An operation analogous to refinement in base Modelleyen (which reduces lists of SVs to their intersection) is needed to identify the \textit{shared part} of two or more SPNs. For this, we define:

\begin{definition}
    An SPN $P_0=[(k_0, SN^0_0),\ (k_1, SN^0_1),$ $...(k_N, SN^0_N)]$ is \textbf{satisfied by} another SPN $P_1=[(k_0, SN^1_0), (k_1, SN^1_1), ... (k_N, SN^1_N)]$ (with the same set of keys  $K=[k_0, k_1, ... k_N]$) given a potentially partial \textit{assignment} $f: V(P_0) \rightarrow V(P_1)$, where $N(P_i)$ is the set of all nodes across all state networks of $P_i$, if and only if the following conditions hold: (1) For $\forall n_0 \in N(P_0)$, $f(n_0)$ is defined (has a mapped node in $N(P_1)$), and (2) For $\forall e_0=(n_0, n_1) \in E(SN^0_i)$ where $E(SN^0_i)$ is the set of all edges in state network $SN^0_i$ in $P_0$, there exists a path in $SN^1_i$ from $f(n_0)$ to $f(n_1)$.
\end{definition}

Intuitively, $P_0$ is satisfied by $P_1$ under an assignment if every node in $P_0$ has a corresponding target node in $P_1$, and every edge in $P_0$ has a path in $P_1$ connecting the assigned targets of its endpoints within the same SN. This ensures that all entities and relations in $P_0$ are present in $P_1$, even if mediated by additional entities not in $P_0$ (as paths, not direct edges, are required).

We can now redefine "finding the intersection" of two SPNs $P_0$ and $P_1$ as "minimally refining $P_0$ to be satisfied by $P_1$." This is achieved through \textit{network refinement with rerelation}, where $P_0$ (source) is refined by $P_1$ (refiner). The process, detailed in Alg. \ref{alg:netref} in the Appendix, relies on two subprocesses:

\textit{Refinement:} Nodes in $P_0$ that are missing in $P_1$, and edges in SNs of $P_0$ that don't have a path between their endpoints in the corresponding SN of $P_1$, are removed.

\textit{Rerelation:} When an edge $(n_0, n_1)$ is removed (including via node removal), a new edge $(p_i, s_i)$ is created for $\forall p_i \in P(n_0), s_i \in S(n_1)$, where $P(n)$ and $S(n)$ are predecessors and successors of $n$ respectively. (Each edge formed by rerelation is also checked for the same conditions of presence.)

\begin{figure}
     \centering
     \begin{subfigure}{0.2\textwidth}
         \centering
         \includegraphics[width=\textwidth]{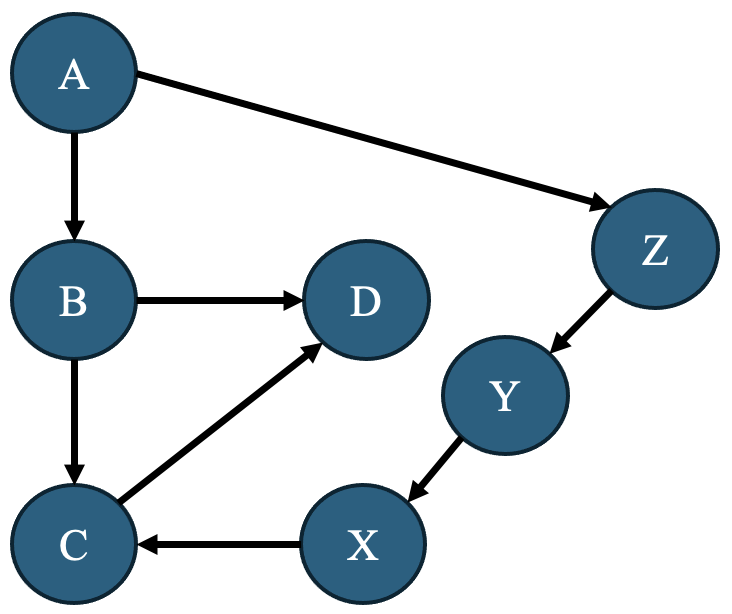}
         \caption{Source SN.}
         \label{fig:netref_source}
     \end{subfigure}
     \begin{subfigure}{0.2\textwidth}
         \centering
         \includegraphics[width=\textwidth]{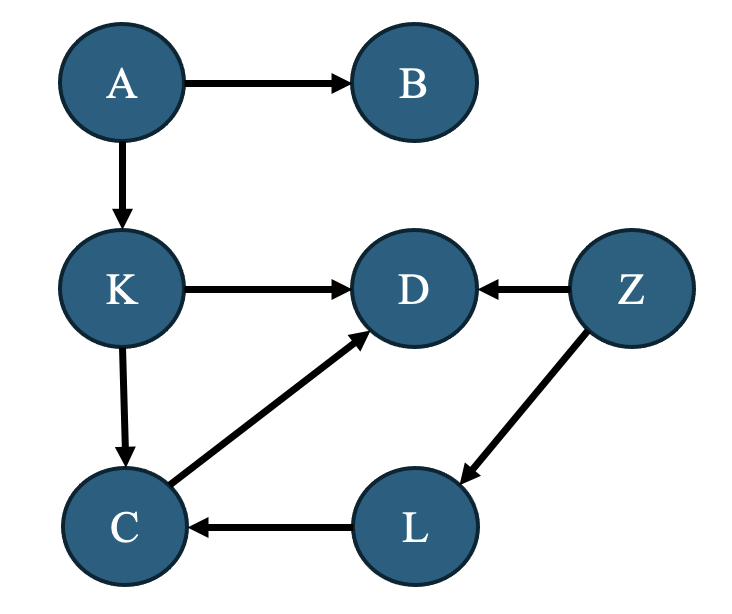}
         \caption{Refiner SN.}
         \label{fig:netref_refiner}
     \end{subfigure}
     \begin{subfigure}{0.2\textwidth}
         \centering
         \includegraphics[width=\textwidth]{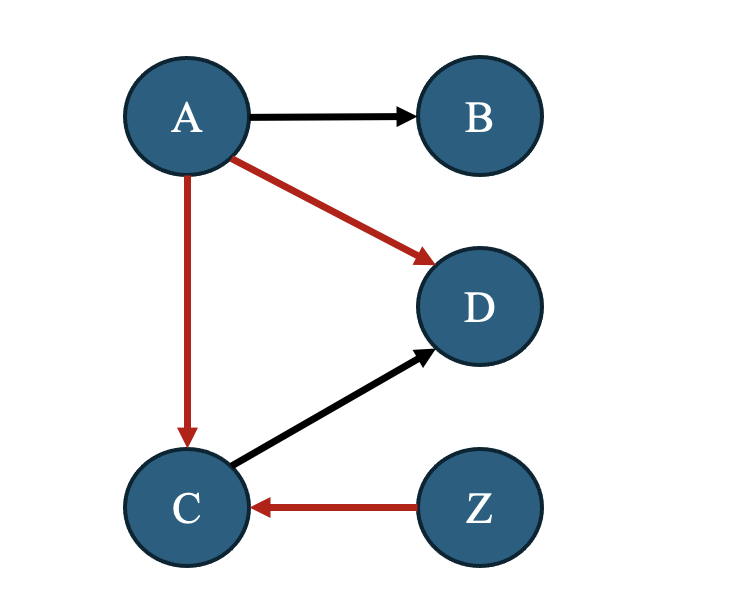}
         \caption{Source, refined.}
         \label{fig:netref_final}
     \end{subfigure}
    \caption{Illustration of network refinement with rerelation. In (c), highlighted edges are created through rerelation. Paths $(A,D)$ and $(A,C)$ exist in both networks but are mediated by different intermediaries (B and K respectively), leading to refined intermediaries and new edges. Similarly, path $(Z,C)$, mediated by $(Y,X)$ in the source and $(L)$ in the refiner, is refined. Edge $(A,Z)$ is removed as it lacks a corresponding path in the refiner SPN. Edge $(A,B)$ is preserved unchanged, as it appears in both networks, despite differing successors of B. (Node positions are illustrative and irrelevant to refinement.)}
    \label{fig:netref}
\end{figure}

Figure \ref{fig:netref} illustrates this process: the source SPN in \ref{fig:netref_source} is refined by the refiner in \ref{fig:netref_refiner}, resulting in the refined SPN in \ref{fig:netref_final}. Paths like $(A,D)$ or $(A,C)$ are preserved despite differing intermediaries. Applying this process sequentially to a source SPN across multiple refiners results in a final SPN representing commonalities across all refiners, ensuring it is satisfied by each refiner retrospectively.\footnote{The local continual learning guarantee in base Modelleyen (Theorem 1) also applies to Algorithm \ref{alg:netref}, as nodes and paths in an SPN are analogous to state variables in the base version. Thus, the theorem's proof holds for node and path removal, with edge refinement being more constrained than path refinement.}

Above, we outlined the key modifications to the base Modelleyen framework to enable its operation on networks. Additional implementation details, excluded from the main text, are provided in Appendix \ref{sec:mnr_details}. These include a statistical refinement mechanism to mitigate noise, a method for identifying proper node mappings, the feature representation used to convert images into networks, and adjustments to the original Modelleyen learning flow to accommodate network representations. One point to note is that the feature representation used for network construction has limitations in representational expressivity, which is largely responsible for the suboptimal accuracy of MNR. Further details are in Section \ref{sec:feature_representation}, with a broader discussion in the Conclusion.

\section{Experimental setup}

\textbf{\textit{Modelleyen}} We demonstrate the operation of Modelleyen on a simple test environment, which is a finite-state machine (FSM), designed to model various types of temporal successions, whose details we omit here due to spatial limitations and present in Appendix \ref{sec:appendix_expsetup}. The agent learns the environment through Modelleyen and selects actions using a simple planner (Sec. \ref{sec:planner}, Appendix) based on the learned model. The environment includes three subtypes ("RS", "SG", "NEG"), and two variants of \textit{vanilla} (regular design) and \textit{random} (with two additional states activating randomly, unrelated to the task, to test the ability to handle noise). This environment was chosen to validate the core operation in a simple and understandable setting, which made in-depth analysis and debug of the design very feasible during development process. There is no inherent limitation to applying Modelleyen to more complex environments, like those used for RL algorithms. However, the planner would need adjustments to make the search nonexhaustive, and an integration of environment modeling with complex visual space processing, such as the MNR variant, is needed. We leave validation on such environments and changes in design to future work, as this presentation is dense enough already. We measure the average steps to reach the goal by an agent that chooses actions using the planner that operates on the environment model that has been learned (with a 10\% chance of random actions for exploration) vs. an agent that acts purely randomly. We conduct continual learning experiments where the agent learns with predefined goals and the environment subtypes switch every 500 steps (with readaptation) or 1000 steps (without readaptation). We test whether the agent can achieve similar performance in different subtypes, both in vanilla and random environment variants without any readaptation of the model, and also analyse learning progression when readaptation is enabled. For more details, see Appendix \ref{sec:appendix_expsetup}.

\textbf{\textit{MNR}} We experiment on MNIST dataset, with the aim to show continual learning performance of MNR, contrast it with the learning progression of neural networks, and investigate learned representations of classes by MNR for proper structure and comprehensibility in some additional results. Our MNR learning process involves randomly selecting $N_C$ classes from the 10 available at the start of each trial. In one \textit{cycle}, the system is exposed to $N_{sample}$ samples from each class sequentially, with one exposure and learning step per sample (no reexposure, as repeated steps have no effect in MNR). Processing samples from one class constitutes an \textit{iteration}. Neural networks follow the same flow, trained on one sample per step until convergence or max epochs. Performance is evaluated via per-class accuracy after each iteration ($N_C$ evaluations per cycle). We train for 10 such cycles, simulating a general and realistic continual flow on information with the requirement of ongoing learning without any constraining assumptions. We run our experiments on MNR with $N_C=3,\ 5$ and $10$, and with $N_C=3$ on NNs for comparison of behavior. Reported results are averages of 10 and 5 runs for $N_C=3$ and $5$ respectively.  Further details, including choice of parameters and computation of predictions in MNR, are on Appendix \ref{sec:exp_details}.

We don't provide comparisons with existing continual learning methods, as none offer a meaningful comparison. As noted in Section \ref{sec:related_work}, all methods we know of rely on restrictive assumptions, such as replay buffers or clear task boundaries, which our approach is specifically designed to avoid. Existing methods can achieve near-perfect information retention under these conditions \cite{erden2024directed}, hence comparing them as baselines wouldn't be informative. Additionally, none of these methods integrate with precise, goal-directed behavior in learned models, making them unsuitable for our behavioral experiments.

\section{Results and Discussion}
\label{sec:results}

\begin{figure}[t]{}
     \centering
     \includegraphics[width=0.5\textwidth]{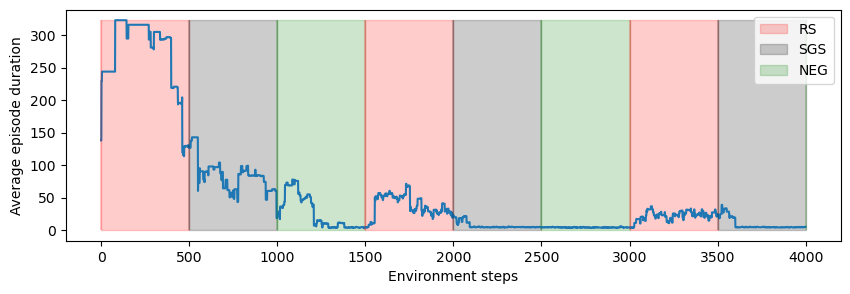}
     \caption{Average (5 trials) episode durations throughout learning with changing environment subtypes, with model readaptation enabled. Vertical limits show the environment changes, note that the actual step of change slightly varies across trials since end of the ongoing episode is waited.}
     \label{fig:CL_planning_nonrand_readapt}
 \end{figure}

\begin{table*}[]
    \centering
    \caption{Continual learning in environment modelling. Mean episode durations with environment change, for vanilla, random environment, and readaptation variants. Columns represent the successive environment subtypes. Subtypes indexed "L" have model learning enabled, "NL" have it disabled (except for "readaptation" variant, which continues learning throughout the end). All results are averages across 5 trials.}
    \begin{tabular}{c|c|c|c|c|c}
         & RS-L & SGS-L & NEG-L & RS-NL & SGS-NL  \\
        \hline
        Vanilla & 45.58 (25.55) & 5.33 (0.28) & 4.47 (0.22) & 10.38 (1.68)  & 4.3 (0.11)\\
        Random Env. & 190.86 (148.0) & 32.3 (9.93) & 9.87 (3.45) & 121.69 (82.33) & 35.05 (5.42) \\        
        Readaptation & 89.01 (58.72) & 28.19 (21.45) & 6.06 (0.74) & 13.73 (3.45) & 4.71 (0.15) \\       
        \hline
        Random actions & 275.86 & 67.53 & 52.48 & 275.86 & 67.53 \\        
    \end{tabular}
    \label{tab:continual}
\end{table*}

\textit{\textbf{Modelleyen:}} Table \ref{tab:continual} displays the agent's continual learning performance across changing environments. Both vanilla and random variants maintain or even improve their performance after exposure to different environments, often outperforming initial learning periods, without readaptation. For instance, the vanilla version averages 5.33 steps on the SGS variant during learning and 4.3 steps after intermittent exposure to other subtypes. Fig. \ref{fig:CL_planning_nonrand_readapt} also illustrates this, showing that with model adaptation enabled, the agent performs consistently with its previous endpoint performance in the same environment subtype, without any spikes indicating destructive adaptation. Additionally, most steps are spent in the RS variant due to the precise timing requirements of the planner (see Sec. \ref{sec:planner} in Appendix).

\begin{figure*}
     \centering
     \begin{subfigure}{0.32\textwidth}
         \centering
         \includegraphics[width=\textwidth]{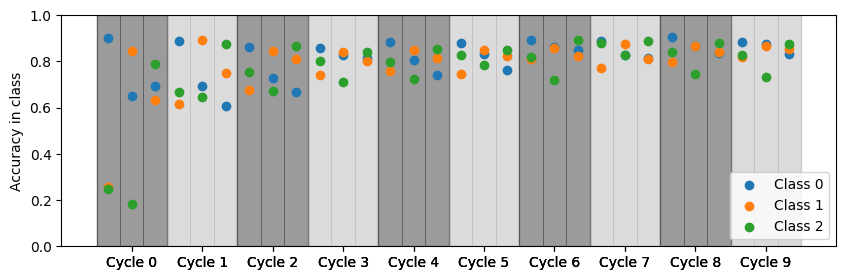}
         \caption{MNR, $N_C=3$.}
         \label{fig:results_cl_my}
     \end{subfigure}
     \begin{subfigure}{0.32\textwidth}
         \centering
         \includegraphics[width=\textwidth]{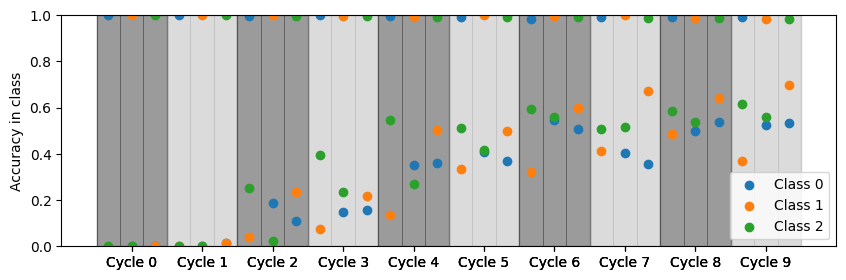}
         \caption{Fully connected NN, $N_C=3$.}
         \label{fig:results_cl_nn}
     \end{subfigure}
     \begin{subfigure}{0.32\textwidth}
         \centering
         \includegraphics[width=\textwidth]{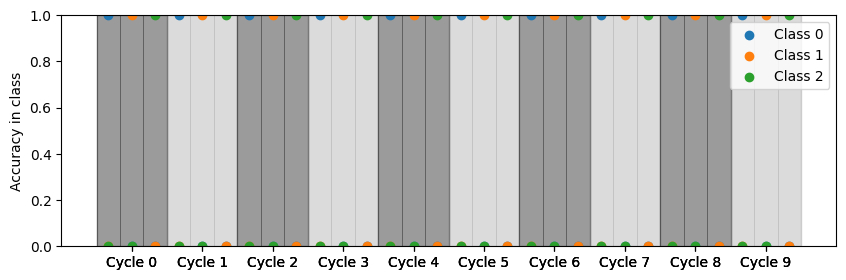}
         \caption{Convolutional NN, $N_C=3$.}
         \label{fig:results_cl_cnn}
     \end{subfigure}
     \begin{subfigure}{0.33\textwidth}
         \centering
         \includegraphics[width=\textwidth]{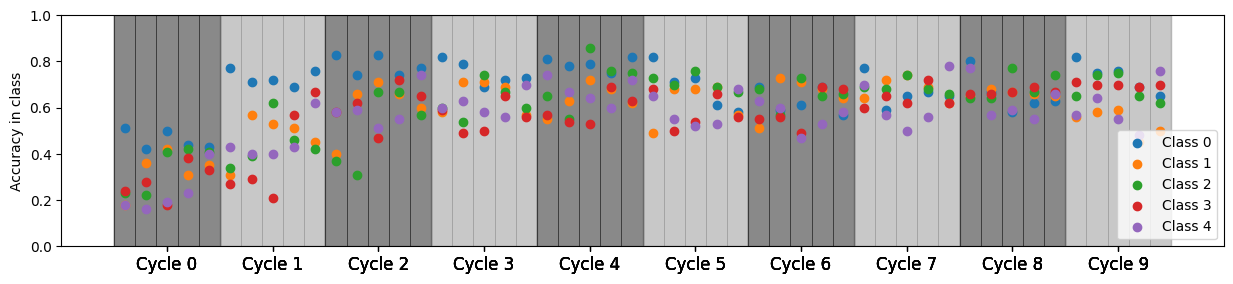}
         \caption{MNR, $N_C=5$.}
         \label{fig:results_cl_my_5c}
     \end{subfigure}
     \begin{subfigure}{0.33\textwidth}
         \centering
         \includegraphics[width=\textwidth]{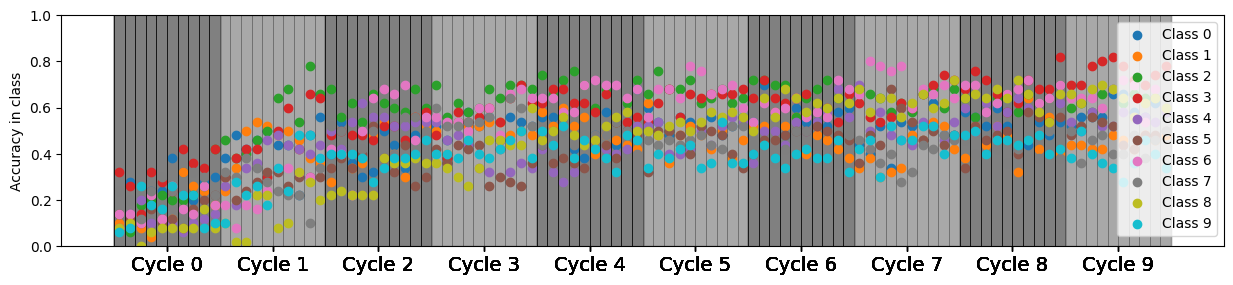}
         \caption{MNR, $N_C=10$.}
         \label{fig:results_cl_my_10c}
     \end{subfigure}
    \caption{Learning performance of MNR, fully connected, and convolutional neural networks on $N_C$-class incremental learning over 10 cycles. Accuracies reflect correct classification ratios for each class. Shaded areas denote cycles, and vertical lines separate iterations within cycles. Results are averaged over 10 (a-c) and 5 (d-e) runs. Note that class indices $i$ are randomly chosen at the start of each run and do \textit{not} necessarily correspond to digit $i$.}
    \label{fig:results_cl}
\end{figure*}

\textbf{\textit{MNR}} Figure \ref{fig:results_cl} shows learning progression of MNR for $N_C=3$, $4$, and $5$ classes; as well as that of the neural network variants for $N_C=3$ for comparison. MNR's final performance, as expected, does not achieve perfect identification, with accuracies of ~85\%, 60\% and 50\% for $N_C=3, 5$ and $10$ respectively after 10 cycles. This stems primarily from limitations of feature representation (Section \ref{sec:feature_representation}) and while it suggests the need for improvement with better representations (see Conclusions), it is not our main focus here. To validate continual learning of our design, we focus on MNR's high retention of learned information, as shown in Figures \ref{fig:results_cl_my}, \ref{fig:results_cl_my_5c}, and \ref{fig:results_cl_my_10c}. Performance on class $i$ remains stable in later iterations ($j>i$) of the same cycle, with early accuracy persisting throughout. This contrasts sharply with neural networks: a fully connected NN (Fig. \ref{fig:results_cl_nn}) loses all information on class $j>i$ in early cycles, and even after 10 cycles, it fails to retain a stable representation, showing $>50\%$ accuracy loss. A convolutional NN performs worse, losing all information repeatedly. Continual learning is most critical in early cycles (first 3-4), as in the long run, with increasing number of cycles, the problem is equivalent to stochastic gradient descent with a slow timescale, reducing the problem to statistical learning with data abundance. MNR's retention of performance is consistent across tests with 5 and 10 classes as well, albeit with lower baseline accuracies. We note that in MNR, as in Modelleyen, even when there are small performance fluctuations, it is by design not due to direct destruction of existing information (unless a conditioner is removed for cumulative insignificance) but stems from over-refinement or negative conditioning.

We briefly note the factors limiting performance of MNR compared to the perfect detection achieved by neural networks. First, the representation used lacks full expressivity, capturing only gradient change points rather than all shape features (see Appendix \ref{sec:feature_representation}). Second, the current statistical refinement (Appendix \ref{sec:mnr_details}) approach retains some features not present in every sample of a class, yet SPN satisfaction (Definition 5) requires precise matches. This leads to missed instances, especially outliers. This is rather straightforward to offset by allowing soft satisfaction of SPNs. These limitations were not the focus of this work, which prioritized validating the learning flow with a demonstrative representation, and will be addressed in future research.

We also note that the internal representations using MNR are inherently comprehensible to human eye. We omit this analysis here, but they can be found in Appendix \ref{sec:mnr_results_comprehensibility}.

\section{Conclusions}

In this work, we analyzed the ability of computational units in a learning system to preserve past responses while adapting to new inputs—a crucial prerequisite for true continual learning. We demonstrated that standard neural networks inherently fail to meet this requirement and introduced Modelleyen, a novel learning algorithm capable of retaining past responses, as verified both theoretically and experimentally. We extended the principles of Modelleyen to network structures, enabling it to process large observation spaces, such as visual data, more effectively. Although the current implementation has limitations, requiring considerable future development to achieve computational efficiency and match modern neural networks in performance; both our theoretical inquiry and the initial experimental results here demonstrate that such an approach is essential to realize continual learning in AI systems, and approaches based on straightforward statistical parameter tuning like NNs are inherently inadequate for learning without loss of past knowledge. This can be more regarded as a class of architectures based on local variation and selection, whose exploration by community is long overdue. To the best of our knowledge, our work is the first inquiry into such a mechanism for learning.

\textit{\textbf{Limitations and future work}} The current design is preliminary and faces several limitations, particularly in its MNR variant for networked observation spaces. Firstly, its implementation is computationally demanding due to redundant modeling introduced for simplicity, as optimization was not prioritized (see Appendix \ref{sec:mnr_details}). Upstream CSVs are represented as large networks instead of distinct subnetworks, with each CSV limited to one target, inflating computation cost. Also, learning flow processes multiple intersecting upstream paths simultaneously, which could be streamlined by focusing on the "best-matching" path. While Modelleyen allows for complexifying as much as needed, operations should ideally involve a small set of CSVs with combined SPNs no more complex than the SPN of the observation, capping per-step complexity to the observation space size - we aim to realize this goal. Secondly, the current feature representation, based on gradient sign changes, is simple and tailored for 2D shape detection but has limited expressivity (see Appendix \ref{sec:feature_representation}), affecting performance. Future work could refine this method or adopt established approaches like SIFT (\cite{lindeberg2012scale}), pretrained visual models (\cite{oquab2023dinov2}), or frequency components (\cite{xu2020learning}), all compatible with the MNR flow. Pixel-level detection, similar to neural networks, or intermediate representations like CNN-style filters could also improve performance and expand applicability, though scalability and interpretability must be considered. Additionally, Modelleyen is currently explored separately for environment dynamics modeling and behavior. A natural future direction is modeling the dynamics of environments observed as networks (i.e., network dynamics), enabling experiments with complex visual spaces. Similarly, temporal dynamics, particularly in non-Markovian environments with long-term dependencies, can also be effectively represented, processed, and learned as networks using our refinement algorithm.


\newpage

\appendix

\section{Appendix}

\subsection{Planner}
\label{sec:planner}

Here we describe our planner, an extension on Modelleyen designed to demonstrate goal-directed planning through backward tracking from desired goal states to current states.

\textbf{\textit{Preprocessing the model and Group SVs:}} We first briefly preprocess a learned model to reduce the number of connections. To this end, we group the sets of BSVs in our that are either (1) collectively act as positive or negative source of a CSV, or (2) have an event that is collectively predicted by a CSV. Each such grouping becomes a \textit{constituent} of a Group SV (GSV). For example, if a CSV $C0$ has positive sources $(B0, B1, B2)$ and predicts deactivation of $(B3, B4)$; then two GSVs are created: $G0 = (B0, B1, B2)$, $G2 = (B3, B4)$. This preprocessing stage is only for practical purposes and is not in principle needed for the operation of the planner, but we think it is essential for scalable representations of models learned by Modelleyen in the long run.

\begin{figure*}
     \centering
     \begin{subfigure}[t]{0.25\textwidth}
         \centering
         \includegraphics[width=\textwidth]{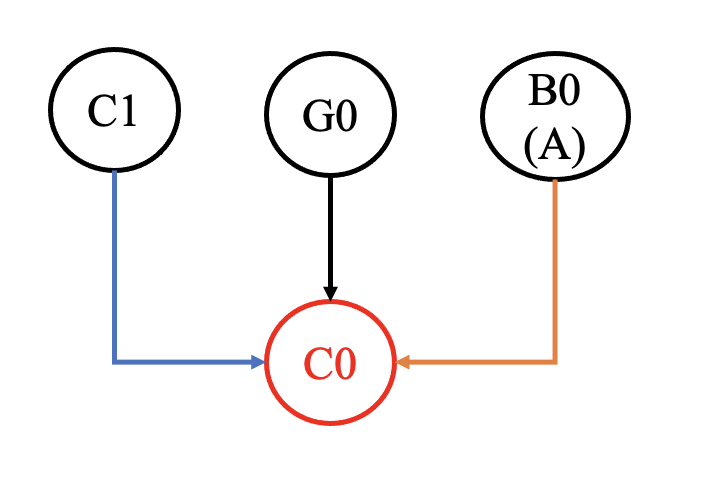}
         \caption{}
         \label{fig:angen1}
     \end{subfigure}
     \vline
     \begin{subfigure}[t]{0.5\textwidth}
         \centering
         \includegraphics[width=\textwidth]{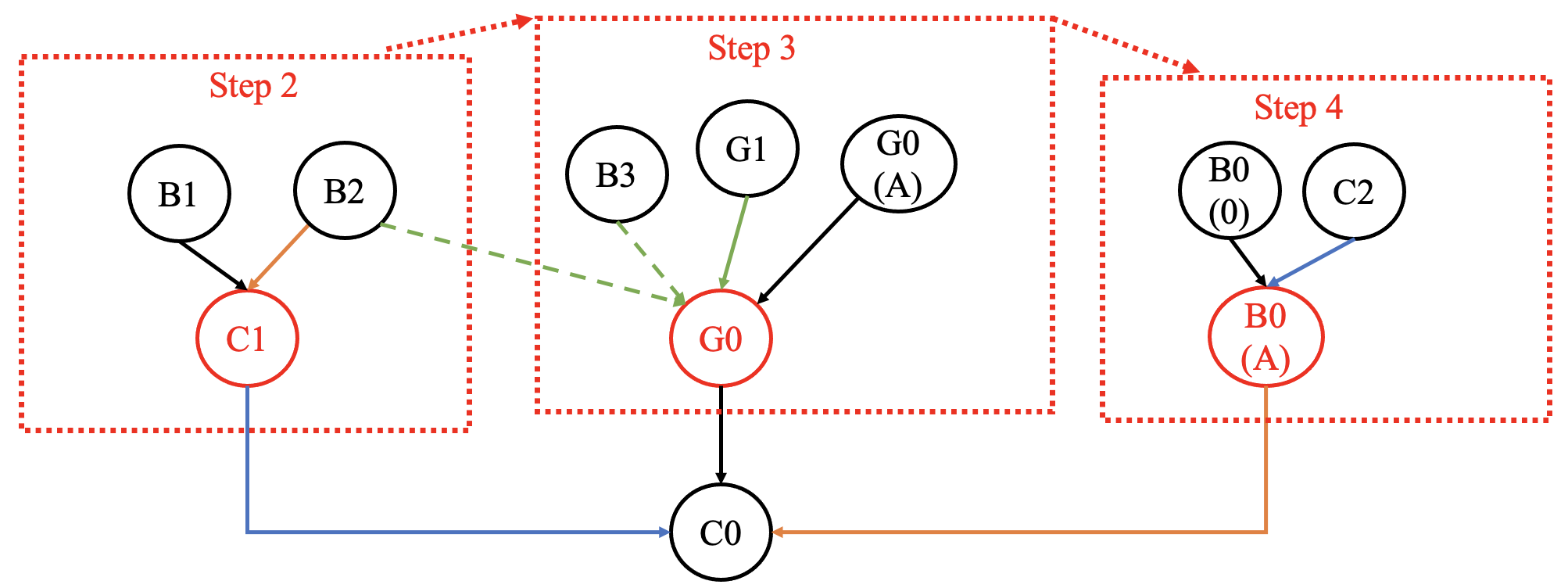}
         \caption{}
         \label{fig:angen2}
     \end{subfigure}
    \caption{Illustration step-by-step upstream generation of action network, operating on different SV types. BX, CX and GX stand for BSV, CSV and GSV nodes respectively, (A) for activation, (0) for nonactive state. Black arrows are positive sources and precondition targets, green arrows are constituent (dashed) and constituency (solid) relations. The node that is extended at each step is highlighted in red. (a) Step 1. CSV C0 is opened. For CSVs, their upstream conditioners (C1) and sources are expanded (G0, B0(A)). (b) Steps 2-4. Each step opens up one of the sources of previous step. For GSVs (G0), constituents (B2, B3), constituencies (G1) and precondition events (G0(A)) are opened. For DSVs (B0(A)), their precondition states (B0(0)) and their conditioners (C2) are opened. Possible interrelations (e.g. B2 for C1, G0) do not need reopening if they already exist.}
    \label{fig:angen}
\end{figure*}

\textbf{\textit{Main Process of the Planner:}} The planner constructs an action network (AN) based on a model generated by Modelleyen, incorporating alternative outcomes. An AN is a dependency graph with root nodes representing the current environmental states (current BSV, GSV, and DSVs), along with possible alternative connections (shown by multiple conditioning links from CSVs) needed to achieve a specified goal state variable. To build this, we use a simple recursive function that generates the upstream action network for a given node (Figure \ref{fig:angen} - see Algorithm \ref{alg:planner} for details). At each call, the function adds predecessors for the specified node until it reaches the root nodes that represent current environmental states. These predecessors vary by state variable types based on their model functionality, as summarized in Figure \ref{fig:angen2}.

\textbf{\textit{Action Choice:}} The agent generates an action network each time it needs to select an action. (While this is computationally unnecessary—since the agent could reuse a generated AN until it reaches the goal by tracking its position along the AN—we maintain this approach for simplicity.) From the generated AN, the agent identifies actions that can immediately activate any CSV in the action model, specifically those whose sources and sources of their downstream targets do not involve any unactualized BSV states. The agent then randomly selects one of these actions for the current step. Since only one action is chosen, the agent can consider the entire AN including alternative pathways.

This planner is explicitly goal-directed, identifying a path from initial states to the goal without needing rewards, although rewards can help prioritize the search. Unlike methods like model-based RL, which typically search from initial states to goals via forward-sampling, the planner considers both initial and goal states, focusing on steps derived from the environment model. The planning algorithm is a simple search method that unfolds upstream action networks from the model, as our main aim is to demonstrate the interface between Modelleyen’s modeling components and general deliberative behavior without going into extensive detail. Planning is a well-established field with efficient methods and useful heuristics \cite{ghallab2016automated}, and once the interface between Modelleyen and the planner is established, implementing more advanced algorithms is straightforward.

Finally, we note two visible limitations of the current version of the planner. First, the generated action networks are exhaustive, including every possible path to initial states. Second, the current version does not account for the precise timing of multiple events. In our experiments, for instance, the RS environment subtype (see Figure \ref{fig:environment}) takes longer due to the BSV \textit{DO} having two pathways for deactivation, the correct one being the one that deactivates BSV \textit{W} as well at the same time. The planner fails to distinguish between these pathways, leading to some unnecessary loops. These limitations are not addressed in current framework to keep its simplicity, since they do not affect our demonstrative use of the planner to a major degree. They are discussed in the Conclusions of the main text.

\paragraph{Overview of the Agent's Operation Flow} In summary, the operation of an agent utilizing Modelleyen and this planner follows these steps, repeated continuously as the agent interacts with the environment in an online manner, without the need for episode division or offline learning periods:

\begin{enumerate}
    \item Execute actions and gather the resulting observations from the environment.
    \item Process the environment's observations and update the model (Modelleyen - Algorithms \ref{alg:algorithm_adaptationloop} and \ref{alg:algorithm_csvstate}.)
    \item Generate a plan based on the current model and goals, then select an action from the resulting plan (Planner - Algorithm \ref{alg:planner}.)
\end{enumerate}

\subsection{Details of Modelleyen}

\subsubsection{Details of system components}
\label{sec:modelleyen_details}

\begin{figure}[t]{}
     \centering
     \includegraphics[width=0.2\textwidth]{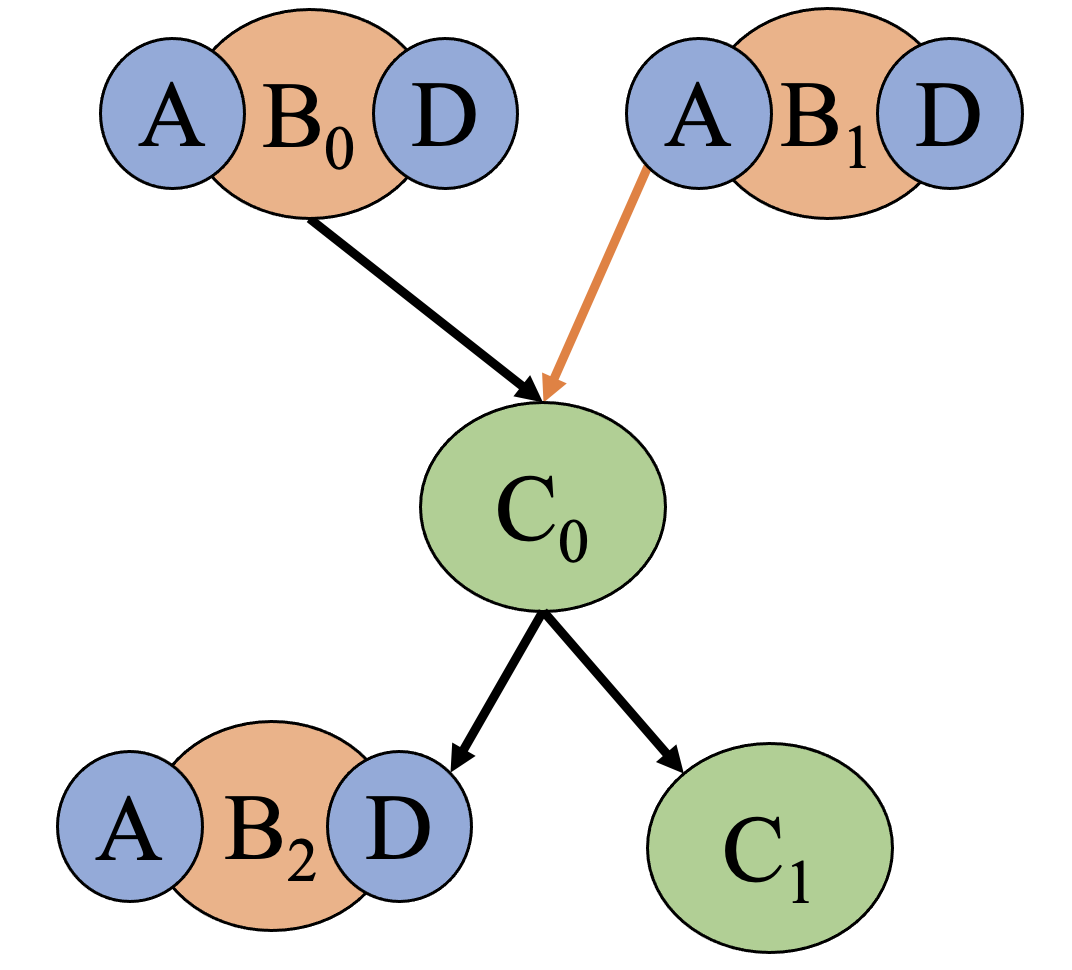}
     \caption{Illustration of SV types and relationships. The figure shows BSVs ($B_i$), their DSVs for activation (A) and deactivation (D), and CSVs ($C_i$). Here, CSV $C_0$ takes as positive source BSV $B_0$, as negative source the activation DSV of $B_1$; and conditions the CSV $C_1$ as well as the deactivation of $B_2$, modelling "$B_2$ is deactivated and $C_1$ is active if $B_0$ is active and $B_1$ is not activated."}
     \label{fig:svs}
 \end{figure}

We define a state variable (SV) as a variable that can take three values: 1 for \textit{active}, -1 for \textit{inactive}, and 0 which can be interpreted as \textit{unobserved, undefined,} or \textit{irrelevant} depending on context. Note that the numerical values are given only as shorthand notation and do not participate in an algebraic operation anywhere. The phrase \textit{nonactive} refers to any SV that is not active. The SV construct comes in three subtypes: Base SVs (BSVs), Dynamics SVs (DSVs), Conditioning SVs (CSVs).

\textit{BSV:} BSVs are the externally-specified SVs whose states, which is assumed to be either 1 or -1, are provided externally to the system at each time instant. These can be regarded as the direct observations from the environment.

\textit{DSV:} Each BSV comes with two associated DSVs, for activation (A-DSV) and deactivation (D-DSV) respectively. Activation at timestep $t$ is defined as the transition of a BSV state from -1 in step $t-1$ to 1 in step $t$; and likewise deactivation at $t$ is defined from 1 in $t-1$ to -1 in $t$. At step $t$, A-DSV is deduced active (state 1) if activation is observed at step $t$, inactive (-1) if a BSV is inactive at $t-1$ and no activation is observed at $t$, and undefined (0) if the BSV is already active. Symmetrically, at step $t$, D-DSV is deduced active (state 1) if deactivation is observed at step $t$, inactive (-1) if a BSV is active at $t-1$ and no deactivation is observed at $t$, and undefined (0) if the BSV is already inactive. The BSVs are modelled only through changes in their states via their associated DSVs, and are not predicted by themselves.

\textit{CSV:} A CSV is a SV that conditions either DSVs or other CSVs (but not BSVs since they are not subject to direct modelling of their states); that is, predicts their activation. More specifically; each CSV comes with a set of positive and negative sources, where each source is either a BSV or DSV; and a set of targets, which correspond to the SVs that this CSV conditions. At steady state, a CSV’s source conditions are said to be satisfied when all its positive sources were active and all its negative sources were nonactive in the previous step - in other words, the satisfaction corresponds to the condition $all(positive\ sources)\ and\ not(any(negative\ source))$ in the previous step. A CSV state is undefined (0) if its source conditions are not satisfied. If its source conditions are satisfied; a CSV’s state is active (1) if the state of all its targets are either active or unobserved; and inactive (-1) if the state of all its targets are either inactive or unobserved. In case inactive and active targets are observed together, the CSV is duplicated to encompass the corresponding subsets of targets (as detailed below), hence we always ensure that one of the two above conditions will be satisfied with respect to the states of the targets. A CSV is to be interpreted as a state variable that represents the observance of a particular relationship - it being active means that this particular relationship (e.g. a change, as represented by a DSV, is observed conditioned on some sources) is observed, and it being inactive means that this relationship is not observed. The CSV being undefined or unobserved corresponds to the case in which the conditions for the observation of the relationship are not satisfied in the first place.

Potential targets of conditioning (i.e. DSVs and CSVs), when they are not undefined, are expected to be active if one of their conditioners are active; and inactive otherwise. Furthermore, these types of SVs also possess an \textit{unconditionally} flag, that allow for exceptions in this activity prediction, and are used to model uncertainty regarding activation of SVs. This flag can take three values: It starts with a value "unconditional" at the creation of the CSV and, if the CSV is observed to always be active whenever its sources were satisfied, it remains so. At the first observation of a case where the sources of the CSV are satisfied without the CSV being active, this flag changes to "conditional," signalling that sources alone do not suffice for the activation of the CSV and activity of one of its upstream conditioners is expected. The "conditional" value persists until the first observation of a case where CSV is observed active without any upstream conditioner being active and no new conditioner could be formed (see below and the main text); in which case the flag changes to "possibly unconditional" and remains as such.

Over the course of interaction with the environment, Modelleyen learns a model that predicts the BSV states at the next step indirectly via the prediction of the DSV states. Within the predictions uncertainty is also represented where needed, as apparent from the description of the SVs. Since uncertainty is represented in a local basis (by unconditionality flags of individual SVs), and since CSVs are points of connection relating potentially multiple sources to potentially multiple targets; the uncertainty representation can represent alternative correlated outcomes in a tree-like manner where each downstream “branch” corresponding to the alternative outcomes in one direction or another can include multiple outcomes that occur together - we note that representation of uncertainty as such is not possible in a local manner with e.g. classical neural networks.

\subsubsection{Learning the model}
\label{sec:modelleyen_details_learning}

First, we provide an overview of the learning process in one step of interaction with the environment. During a step, the model is traversed, and the states of all its SVs are computed. For CSVs sources and targets are modified to be able to match the current states to the predictions/explanations of the CSV, so that the model is consistent with the environment at each step. After that, new CSVs are generated for the DSVs and CSVs that lack an explanation at the current step. The new CSV takes as positive sources all currently active eligible SVs in an exhaustive manner. Finally, model is refined by removal of unnecessary state variables.

The learning process is summarized formally on Algorithms \ref{alg:algorithm_adaptationloop} and \ref{alg:algorithm_csvstate}. Below, we provide a detailed breakdown of the processes described on those algorithms.

\begin{algorithm}[tb]
\caption{Pseudocode of the main Modelleyen adaptation loop; formed of state computations followed by CSV generation for unexplained SVs.}

\label{alg:algorithm_adaptationloop}
\textbf{Parameter}: $N$ Set of all target nodes\\
\textbf{Function} \textit{ProcessEnvironmentStep}(observations)
\begin{algorithmic}[1] 
    \STATE $BSVStates \leftarrow \ observations$
    \STATE $ComputeDSVStates()$ \COMMENT{Computes DSV states by BSV events}
    \FOR{$level\ \in\ reverse(ComputationLevels)$}
        \FOR{$CSV\ \in\ SVs_in(level)$}
            \STATE $ComputeState(CSV)$
        \ENDFOR
    \ENDFOR
    \STATE $UnexplainedSVs \leftarrow [SV:\ SV.state = 1\ \AND\ NoConditionerActive(SV)]$
    \STATE $sources \leftarrow [SV:\ SV in\ [BSVs, DSVs]\ \AND\ SV.state = 1\ \AND\ isEligible(SV)]$
    \STATE $ NewCSV = CreateCSV(sources, [SV: SV\ in\ UnexplainedSVs\ \AND\ TargetEligible(SV)])$
    \STATE $ModelRefinement()$ \COMMENT{Removes CSVs with no source or target}
\end{algorithmic}
\end{algorithm}

\begin{algorithm}[tb]
\caption{Pseudocode for CSV state computation.}

\label{alg:algorithm_csvstate}

\textbf{Function} \textit{ComputeState}($CSV$)
\begin{algorithmic}[1] 
    \IF{$AnySourceActive()$}
        \STATE $SeparateActiveInactiveTargets()$ \COMMENT{Creates two CSVs from current one with active and inactive targets in either of them}
        \IF{$AnyTargetObserved()$}
            \STATE $State = 1$
            \STATE $PosSources \leftarrow [source:\ source\ in\ PosSources\ \AND\ source.state=1]$
            \STATE $NegSources \leftarrow [source:\ source\ in\ NegSources\ \AND\ source.state!=1]$
        \ELSIF{$AnyTargetInactive()$}
            \IF{$not(AllSourcesActive())$}
                \STATE $State = 1$
            \ELSE
                \IF{$AnyNegativeSourceActive()$}
                    \STATE $State = 0$
                    \STATE $NegSources \leftarrow [source:\ source\ in\ NegSources\ \AND\ source.State=1]$
                \ELSE
                    \STATE $State = -1$  \COMMENT{No negative source active to explain inactivity of targets}
                \ENDIF
            \ENDIF
        \ENDIF
    \ELSE
        \STATE $State = 0$ \COMMENT{Unobserved if targets are not observed}
    \ENDIF

    \IF{$State = -1$}
        \IF{$NegativeConnectionsFormed$}
            \STATE $FormNegativeConnections()$
        \ELSE
            \STATE $unconditionality = "isConditional"$ \COMMENT{-1 for }
        \ENDIF
    \ENDIF

\end{algorithmic}

\end{algorithm}

Initially, the model is generated with only BSVs and their associated DSVs, and without any CSV. At every step, the current and previous states of all the SVs are recorded, as well as the current and previous events (activation and deactivation) of every BSV. 

At each step, the effective network created by DSVs and CSVs are traversed in the reverse order of computation, similar to backpropagation algorithm; starting from DSVs, then the CSVs that condition these BSVs, then the conditioners of these CSVs, and so on. Each traversed SV gets their state computed, and additionally CSV compositions are changed where needed, as in Figure \ref{fig:csvform} and detailed below.

\textit{Processing of a CSV}

The process for CSVs are carried as follows: If no positive source of a CSV is observed at a given step, its state is deduced as 0 (undefined/unobserved). If at least one source is observed, and if there are both active and inactive targets among the CSV targets, then the CSV is duplicated with different target sets to create one copy that includes active targets and one copy that includes inactive targets (and any undefined targets are shared by both). This ensures that the CSV remains consistent, since it’s activation represents the activation of all its targets provided they are not undefined. There is no way to say whether an undefined target will be consistent with one duplicate or another after the changes to the CSV described below without observing a non-undefined state in them, so they are put into both copies and do not otherwise affect the state deduction of the CSV (except if all targets are undefined, see below).

Following this operation, if a CSV has any target active, then its state is deduced as active (1). If there is no perfect match with the standing sources of CSV and their activations (i.e. there are either inactive positive sources or active negative sources), these source lists are refined so that the remaining sources correspond perfectly to the current state of the network - in other words, any positive source that is inactive and any negative source that is active is removed. This refinement eliminates parts of the previously-posited relationships “hypothesized” to be necessary by the CSV in an exhaustive manner (see details on CSV formation, below) that are observed to be not necessary for the observation of the effect that the CSV models (Figure \ref{fig:csvform_3}.

If, on the other hand, the CSV has any inactive target (which is exclusive with any target being active due to the duplication-differentiation operation made above) and if not all its positive sources are active, then the state is deduced as 0, being consistent with the interpretation of a CSV as being defined only if all its positive sources are active. If however, all positive sources are active; then we look if any negative source is active that can justify the inactivation of the targets of the CSV. If there is at least one negative source that is active, we deduce the state as 0 since source conditions are not satisfied; and refine the negative targets that are not currently active in the same manner we described in the previous paragraph (due to the observation that they are seen to be not necessary for the suppression of the CSV - Figure \ref{fig:csvform_5}).

If, instead, all the targets of CSV are undefined, then the CSV is undefined as well.

A CSV is always created with only positive sources at first and no negative sources, and a CSV always starts as an unconditional CSV for whom we never expect to observe an inactive state (see below part for details on the generation of CSVs). At the observation of an inactive state in the CSV (i.e. one in which sources are active but targets are inactive), only once after the creation of the CSV, we duplicate the CSV and separate the targets that are currently undefined (to protect them from the change being made). In the duplicate that has the inactive targets, we connect the CSV with the negative sources by forming a negative sources list that encompasses all the currently-active eligible BSVs and DSVs in the model, which will be subject to future refinement (criteria of \textit{eligibility} is detailed below, essentially corresponding to SVs that do not yield useful information). This, essentially, attempts to explain the CSV’s observed inactivation. If, however, an inactive state is observed despite already having formed connection with negative sources, then the unconditionally flag of the CSV is set to "conditional", representing that the CSV’s state is now uncertain (setting aside its possible conditioners).

\textit{CSV generation and model refinement}
\label{subsec:csv_form}

After the traversal of SVs for computation of their states and modifications in CSV compositions, all DSVs and CSVs who are observed active but are neither unconditional nor have an active conditioner that explains their activation are labelled as \textit{unexplained}. We then form a CSV that, as positive sources, has all the eligible, currently-active BSVs and DSVs; and as target, has all the eligible SVs in unexplained list (Figure \ref{fig:csvform_1}). Any target which is left outside of this CSV, and hence remain unexplained, have their unconditionally flags set to "possibly conditional" (which basically signals that the SV can go active without any explanation or predictor).

Finally, at the end of the step, we refine the general model by removing any CSVs that may be duplicates of other CSVs (ending up representing the same thing from different histories), as well as any CSV that has no sources or targets left as a result of refinement or duplication operations.

\textit{Source eligibility for CSVs}

To reduce model complexity and avoid the need for repeated exposures to the environment, we pre-filter sources during CSV formation or CSV negative-sources formation by their eligibility as follows: We define \textit{trivial sources} of a CSV as the sources of all the SVs that lie downstream starting from this CSV (i.e. SVs conditioned by this CSV, and CSVs conditioned by them, and so on), plus the associated BSV if a DSV is reached. Intuitively, these are the sources whose states can be determined by the knowledge that the CSV is active (since a CSV being active means that it’s target will be active as well, which will inform us about the states of its sources), and hence wouldn’t be informative sources for the current CSV as any information conveyed by them will be trivial. When forming a CSV, among all the currently-active BSV and DSVs, we filter those that provide trivial information to all the unexplained SVs (i.e. prospective targets for the generated CSV) out as positive sources, and take only those that do not provide trivial information as source to at least one of them. Furthermore, after this filtering, if there is a prospective target for which all the remaining prospective sources provide trivial information, then this target is not taken as a target of the CSV and hence remains unexplained.

In a similar spirit, when forming negative sources, we filter out all the candidates that provide trivial information for the CSVs. In addition, however, we filter out any upstream positive source (that is, the cumulative list of all positive sources among all upstream CSVs of this CSV, i.e. its conditioners and conditioners of its conditioners, including itself) because we already know (by the definition of the conditioning process) that there was an instance in which this CSV was observed when the SVs in this list of positive conditioners was also observed; and hence these negative sources would be eliminated in exposure with the same instance again.

\textit{Conditioner formation for unconditional CSVs}

Here we note a modification that we do not employ currently, but is possible: Currently we allow no CSVs to condition unconditional CSVs since they are not informative and hence prevent the model from being minimal. However, we note that allowing for conditioners to be formed to unexplained (no active conditioners)  unconditional CSVs as well could result in these CSVs already having some conditioners learned from the previous encounters with the environment in case they ever turn conditional, reducing the required number of interactions for the learning of the full environment model, at the cost of making the model more exhaustive in terms of what is being modelled. This would require two changes: (1) At CSV formation, not excluding the unexplained CSVs that are unconditional; and (2) when refining positive sources, we create a CSV which takes as its initial positive sources that are being removed, and that conditions the CSV whose sources are being refined currently. This way, instead of removing what was observed to be active at previous encounters at which the CSV was active, we push them to an upper level of computation to represent an alternative condition in which the CSV was observed to be active before.

\subsubsection{Proof of Theorem 1}
\label{sec:app_proof}

Let $X_P^i$ and $X_N^i$ be positive and negative sources of $C$ respectively that remains \textit{after} refinements that instance $y_i$ causes. Since we know that $C$ does not undergo negative sources formation, and that $y_0$ comes before $y_1$, we can say that $X_P^1 \subseteq X_P^0$ and $X_N^1 \subseteq X_N^0$ since only refinements are allowed on $X_P$ and $X_N$ sets of $C$ by our definition of operations.

We now analyse the two possible cases with respect to satisfaction of sources:
\begin{itemize}
    \item If, in the original encounter with $y_0$ the sources of $C$ were satisfied, then we had $S_x=1 \forall x \in X_P^0$ and $S_x=1 \forall x \in X_P^0$. Since $X_P^1 \subseteq X_P^0$ and $X_N^1 \subseteq X_N^0$, we will also have $S_x=1\ \forall x \in X_P^1$ and $S_x=1\ \forall x \in X_P^1$ at the new encounter with instance $y_0$. Hence, if sources of $C$ were satisfied in the previous encounter with $y_0$, they will remain satisfied in the new encounter. The value of $S_C$ can be -1 or 1 if and only if sources of $C$ are satisfied; in which case it is exclusively determined by the state of its targets (-1 if targets are inactive and 1 if targets are active). Since the states of targets are determined by $y_0$ and hence is the same across the past and new encounter with $y_0$; if $S_C=1 (-1)$ in the past exposure with $y_0$, then it will be $1(-1)$ in the new exposure as well.
    \item If, in the original encounter with $y_0$ the sources of $C$ were not satisfied (and hence original encounter yielded $S_C=0$), then we either had $S_x \neq 1\ \forall x \in X_P^0$ or $S_x = 1\ \forall x \in X_N^0$ (note that we defined $X_P^i$ and $X_N^i$ as source sets \textit{after} the refinements; and hence we know that in both cases it will be the whole of positive/negative source sets that have the property, and not a subset of them; since the source SVs that were not a part of that subset will have been refined). Since $X_P^1 \subseteq X_P^0$ and $X_N^1 \subseteq X_N^0$, we will also have either $S_x \neq 1\ \forall x \in X_P^1$ (if former) or $S_x = 1\ \forall x \in X_N^1$ (if latter), both of them not satisfying the sources conditions of $C$ (hence the new encounter with $y_0$ also yielding $S_C=0$.
\end{itemize}

Therefore, in all cases, response to $y_0$ remains identical before and after exposure to $y_1$.

\subsubsection{Learning the statistical significance of encountered relations}
\label{sec:statistical_significance}

The base mechanisms of Modelleyen as described in the main text rest on an attempt of prediction of all encountered changes in state variables in the environment, forming an explanatory/predictive relationship between any two observed events in that attempt of full modelling of the environment. Unlike neural networks (or other statistical learning methods), the naive algorithm does not depend on, but also does not naturally incorporate, a method of statistically averaging and filtering learned relationships. Such a means of estimation of statistical significance of learned relationships can be incorporated into the models learned by modelleyen in a straightforward manner into the learned relationships locally, which in turn can be used to filter out non-significant relationships, hence preventing overcomplexification of the model.

Let $C$ be a CSV, and let $T$ be a target SV of that CSV. We define the event \textit{sources satisfied}, $SS(C)$, to be the event where all positive sources of $C$ are active and all negative sources are nonactive. For each target, we define an \textit{observation} of the target $O(T)$ to be when the target is observed (i.e. either active or inactive, state 1 or -1, as defined in the main text) and an \textit{incidence} of the target $I(T)$ to be when the target is active (state 1). We define the event \textit{concurrence} to be the event where both the sources of $C$ are satisfied and there is an indicence of target, $CC(C,T)=SS(C) \wedge I(T)$.

We quantify the statistical significance of a learned relationship between a set of sources of a CSV and one of its targets as the \textit{amount of increase in the probability of the incidence of the target given the satisfaction of the sources of the CSV}. We define \textit{normalized causal effect (NCE)} as the amount of increase in probability of incidence of $T$ that satisfaction of sources of CSV $C$ causes, normalized by the original probability of incidence:

\begin{equation}
    NCE = \frac{P(I(T)|SS(C)) - P(I(T))}{P(I(T))}
\end{equation}

The conditional probability in the nominator can be expanded as:

\begin{equation}
    P(I(T)|SS(C)) = \frac{P(I(T), SS(C))}{P(SS(C))} = \frac{P(CC(C,T))}{P(SS(C))}
\end{equation}

by our definition of concurrence $CC(C,T)$ above. All of the probabilities can be computed by locally tracking of the number of instances that the corresponding events are observed, when the target is observed (i.e. $O(T)=1$). When the target is unobserved/undefined, by extension none of the other events are observed.

A positive NCE means that $SS(C)$ increases probability of $I(T)$ and a negative NCE means that $SS(C)$ decreases it. An NCE of e.g. 2.0 means that $SS(C)$ increases probability of $I(T)$ to 3 times the original probability. Within the context of our modelling mechanism, a negative NCE means that the relationship between sources of $C$ and $T$ has been learned in the wrong direction - actual negative relations learned in proper direction will still result in positive NCE, because the sources of that relation will go within the negative sources of $C$ instead of the positive ones, still in the end resulting in the $SS(C)$. The lower the magnitude of NCE, the less significant the relationship is.

Given NCE values for each relationship, one can set a positive threshold $\epsilon_T$, where NCE values with magnitude below it are regarded as statistically insignificant. $\epsilon_T$ represents the trade-off between complete modelling and model complexity. After that separation of relationships into significant and insignificant ones, one can proceed either with their removal, or simply with blocking further conditioner formation for them to prevent overcomplexification in an attempt to predict a near-random relationship (i.e. to prevent "fitting the noise"). Since our main aim in employing this mechanism is to prevent overcomplexification, and since removal of such insignificant relationships from the model completely would result in their re-learning if the agent is exposed to them again; we opt for the latter option and block further conditioner formation for them.

NCE values may have other utilities for the processes of the agent. An example might be that it can be used in the prioritization of subgoals in the planner (see main text), where more reliable causal relationhips are prioritized over less reliable ones. We do not investigate into such utilities at this stage.

It’s important to note that the statistical estimates are not precise during the transient phase. This is due to the refinement mechanism, which prioritizes structural revisions and adjustments to make a given CSV align with observations where feasible. During this phase, estimates tend to overemphasize significance. However, these transients are brief, and NCEs insignificant CSVs quickly diminish once the refinements are complete and the CSV sources settle into their final form. Furthermore, this final form is typically less constrained, leading to more exposures over time in the same environment. Alternatively, we could eliminate these inaccuracies by resetting recorded statistics after each change to the CSV's composition, though this would increase the time needed for an NCE value to be deemed reliable. We do not use this approach here, as we do not find the temporary bias toward significance in transient SVs to be an issue, but it can be employed where precision has priority over efficiency.

\paragraph{Effect on continual learning} Notice that there is no change (particularly no decay) in NCE if the target is not observed - hence, this measure of statistical significance does not decay (relationship "forgotten") in case of a changed environment in which the new one does not display the co-occurance of the two events (target and CSV sources being satisfied), as long as its target is not observed in isolation as well. If its target is observed in the new environment, two cases may occur:
\begin{enumerate}
    \item $P(I(T))$ is stable. This would be expected in an already-mature model or in environments where there is not much variability in the occurance of individual targets (even if the conditions under which they occur differ). In this case, there is no change in NCE.
    \item $P(I(T))$ changes. In this case, NCE will change according to $P(I(T))$. Note, however, that additional exposure can only mean a more accurate estimate of the true $P(I(T))$ value - any change in $P(I(T))$ hence does not have a detrimental effect, but instead makes the causal effect estimate more reliable in the context of the complete model; provided that the new environment itself does not have a probability of $P(I(T))$ in itself that is non-representative of the general probability, in particularly one that is excessively higher than the general one. This latter possibility (an immature estimate of $P(I(T))$ and an unnaturally high $P(I(T))$ in the new environment) is the only case in which a previously-learned correct relationship can be wrongly destroyed in case of a changing environment. But even such cases would have no long-term ramifications as $P(I(T))$ for any given target $T$ would reach to a reliable estimate after a few cycles of exposures to environments where $T$ is observed.
\end{enumerate}

The current method of computing and filtering based on statistical significance has one drawback, however; and it is that only first-order significance of relations are considered. In other words: If we have a CSV C0 with a target D0, and C0 (possibly unconditional) is conditioned by another CSV C1, then whether C0-D0 relationship will be regarded as significant or not depends only on the observations of sources of C0 and D0; and will \textit{not} consider their dependency on C1. This may result in unnecessary filtering in cases where a said statistical relationship is insignificant in the absence of a particular upstream conditioner, but becomes significant with that - we also see effects of this limitation to some degree in our results in the main text. Resolution of this limitation requires consideration of and conditioning on higher-order conditioners when computing the NCE value, and is left for future work.

\subsection{Details of MNR}
\label{sec:mnr_details}

\textit{\textbf{Formal algorithm for network refinement with rerelation}} Algorithm \ref{alg:netref} presents network refinement with relation process formally.

\begin{algorithm}[tb]
\caption{Network refinement with rerelation.}

\label{alg:netref}
\textbf{Function-}\textit{RefineBy}($P_0$, $P_1$, f)

\textbf{Parameters:} $P_0$, source SPN. $P_1$, refiner SPN. $f$, a partial assignment between nodes in $P_0$ to nodes in $P_1$.

\begin{algorithmic}[1] 
    \FOR{$SN^0_i \in P_0$}
        \FOR{$n \in nodes(SN^0_i)$}                    
            \IF{$f(n)$ not defined}
                \FOR{$(n_0, n_1) \in edges(SN^0_i, n)$}
                    \STATE RemoveWithRerelation($SN^0_i, n_0, n_1)$
                \ENDFOR
                \STATE $SN^0_i$.RemoveNode($n$)
            \ENDIF
        \ENDFOR
        
        \FOR{$(n_0, n_1) \in edges(SN^0_i)$}
            \IF{$path(f(n_0), f(n_1))$ not in $SN^1_i$}
                \STATE RemoveWithRerelation($SN^0_i, n_0, n_1)$
            \ENDIF
        \ENDFOR
    \ENDFOR

\end{algorithmic}

\textbf{Function-}\textit{RemoveWithRerelation}($SN$, $n_0$, $n_1$)

\begin{algorithmic}[1] 
    \FOR{$(p,s)\ \in\ prod(P_{SN}(n_0),\  S_{SN}(n_1))$}
        \STATE $SN$.AddEdge($p$, $s$)
    \ENDFOR
    \STATE $SN$.RemoveEdge($n_0$, $n_1$)
\end{algorithmic}

\end{algorithm}

\textit{\textbf{Statistical Refinement}} Given the noisy experimental domain, we enhanced Algorithm \ref{alg:netref} by incorporating node/edge observation statistics. Instead of removing a node/edge upon its first absence, we remove it only if the ratio of its absences exceeds a threshold $T_{ref} \in (0,1)$. This prevents losing important features potentially missed due to noise or misassignment (see below). Being a more constrained removal condition, this maintains \textit{Modelleyen}'s continual learning guarantees. The inconsistency with past responses in this modification is intentional, as the mechanism is designed to prevent adaptation to infrequently observed instances, which are treated as noise. In the case of statistical refinement, past responses are preserved only for inputs that were observed frequently enough, above the defined insignificance threshold.

\begin{figure}
     \centering
     \begin{subfigure}{0.2\textwidth}
         \centering
         \includegraphics[width=\textwidth]{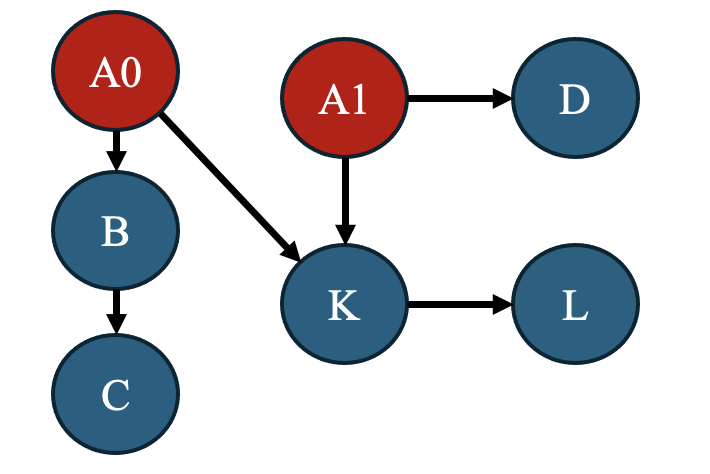}
         \caption{Source SN.}
         \label{fig:assignment_source}
     \end{subfigure}
     \hspace{1cm}
     \begin{subfigure}{0.2\textwidth}
         \centering
         \includegraphics[width=\textwidth]{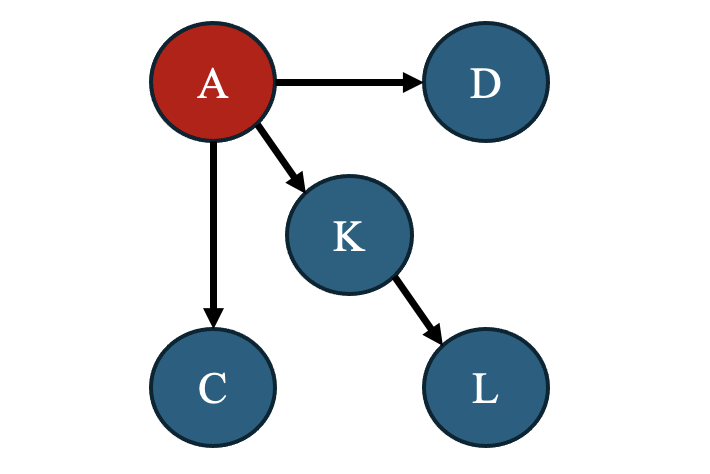}
         \caption{Refiner SN.}
         \label{fig:assignment_refiner}
     \end{subfigure}
     \vfill
     \begin{subfigure}{0.2\textwidth}
         \centering
         \includegraphics[width=\textwidth]{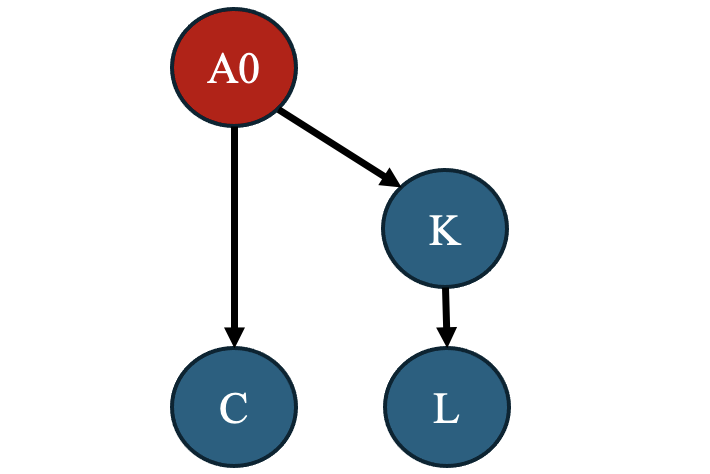}
         \caption{Refinement result with $A$ assigned to $A0$.}
         \label{fig:assignment_final_0}
     \end{subfigure}
     \hspace{1cm}
      \begin{subfigure}{0.2\textwidth}
         \centering
         \includegraphics[width=\textwidth]{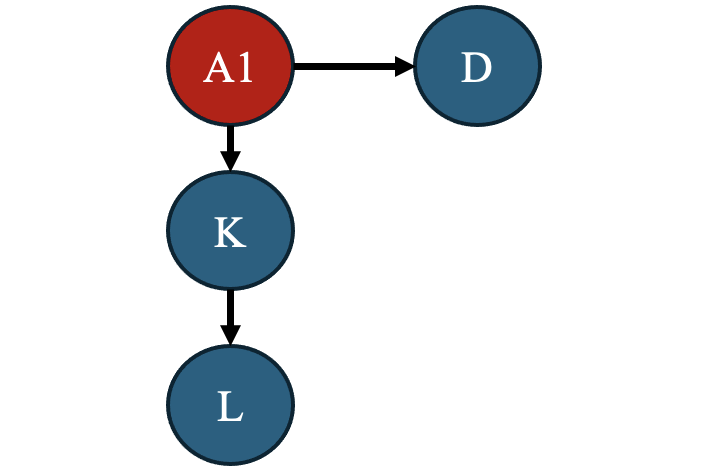}
         \caption{Refinement result with $A$ assigned to $A1$.}
         \label{fig:assignment_final_1}
     \end{subfigure}
    \caption{A node ($A$) of a given type (red) in the refiner SPN can map to multiple nodes in the source network ($A0$ or $A1$), resulting in two post-refinement networks with no clear superior assignment.}
    \label{fig:assignment}
\end{figure}

\textit{\textbf{Assignments}} A key issue in extending the base Modelleyen framework to network refinement is finding a suitable (possibly partial) node mapping $f: V(P_0) \rightarrow V(P_1)$) between two SPNs. The only constraint is that source nodes can only map to nodes of the same type (i.e., representing the same feature). However, multiple valid mappings may exist, resulting in different post-refinement structures (see Figure \ref{fig:assignment}). To address this, we used the following approach: We create a population of alternative assignments by pairing nodes from the source and refiner SPNs based on shared node types. While this works well with a small number of nodes per type, our feature representation (Section \ref{sec:feature_representation}) often requires additional prioritization. To address this, we rank node pairs by their positional proximity in both SPNs, using the negative softmax of the distance between candidate pairs to probabilistically select the most suitable assignments. After generating the population, we calculate a \textit{mismatch score} for each assignment, representing the total number of nodes and paths in the source SPN missing from the refiner SPN. The assignment with the lowest mismatch score—requiring the least refinement—is selected.

In our experiments, this assignment mechanism produced effective mappings (see Section \ref{sec:results}). However, it is not flawless, and selecting the optimal assignment from the population remains the most computationally intensive step in our workflow. Further research could likely improve this process or even eliminate the need for such assignments altogether.

\subsubsection{Changes in learning flow}
\label{sec:learning_flow}

As in base Modelleyen, our system design defines a \textit{model} using \textit{conditioning state variables (CSVs)}, which describe relationships between their \textit{sources} (state polynetworks, or SPNs) and \textit{targets} (other CSVs or specific target state variables). The learning flow largely follows Modelleyen's approach, except for the following adjustments:

\begin{enumerate}

    \item Unlike base Modelleyen, where upstream CSVs integrate with lower-level CSVs via an \textit{and} condition, our implementation treats upstream CSVs as observed \textit{subvariants} of their CSV targets, with source SPNs encompassing the source SPNs of them. This allows assignments from lower-order CSVs to propagate upstream, eliminating redundant assignments and simplifying the learning flow by avoiding the need for additional subnetwork definitions.

    \item Unlike base Modelleyen, which creates a common CSV for all targets in a step, we assume a single target per CSV and create separate CSVs (with identical sources) for each target. This change supports the upstream assignment propagation in previous point.

    \item Instead of embedding negative (suppressing) sources within a CSV, we externalize them into separate CSVs. A negatively-conditioning CSV is formed when a state variable with an inactive state and no active negative conditioner is observed (with potentially multiple formed per target). We also redefine the \textit{unconditionality} flag to deactivate upon the first observation of an inactive state, allowing simultaneous positive and negative upstream conditioning.

    \item Instead of the complex $NCC$ metric, we filter insignificant conditioners by removing a conditioner $C$ of target $T$ if $P(SS(C)|I(T)) < T_{sign}$, where $SS(C)$ and $I(T)$ represent the satisfaction of $C$'s sources and the observation of $T$'s state, respectively.

\end{enumerate}

The major changes, outlined in points 1 and 2, primarily affect CSV composition without altering the core learning flow or CSV state definitions. These changes were made for implementation simplicity and will be modified in future framework developments to include multiple targets and distinct upstream networks, improving representational efficiency (see Section \ref{sec:results}).

Currently, the operational flow is computationally intensive, limiting simulations to shorter durations. At this stage, our focus is on designing and validating the learning flow rather than optimizing the algorithm or capping model complexity, though these aspects are briefly discussed in the Conclusions of the main text.

\subsubsection{Feature representation for basic shape learning}
\label{sec:feature_representation}

\textit{Modelleyen with network refinement} (MNR) is applicable to any observation space representable as networks. For our experiments, we focus on demonstrating the method in a basic visual processing domain: 2D shape identification in binary images, using MNIST as the test domain. This task is foundational in computer vision and has historically served as a starting point for approaches like neural networks \cite{lecun1998gradient, cortes1995support}. Below, we detail the feature representation (image-to-network conversion) used. We note that this is only demonstrative for the use of our approach in a simple context, but future work can extend it to other types, including 2D features like color gradients or 3D features with spatial positions, as the domain requires.

\begin{figure}
     \centering
     \begin{subfigure}[t]{0.2\textwidth}
         \centering
         \includegraphics[width=\textwidth]{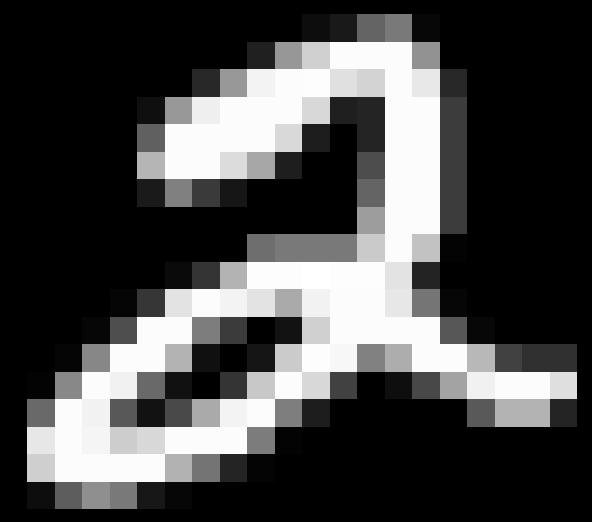}
         \caption{Original image.}
         \label{fig:representation_example_im}
     \end{subfigure}
     \hfill
     \begin{subfigure}[t]{0.2\textwidth}
         \centering
         \includegraphics[width=\textwidth]{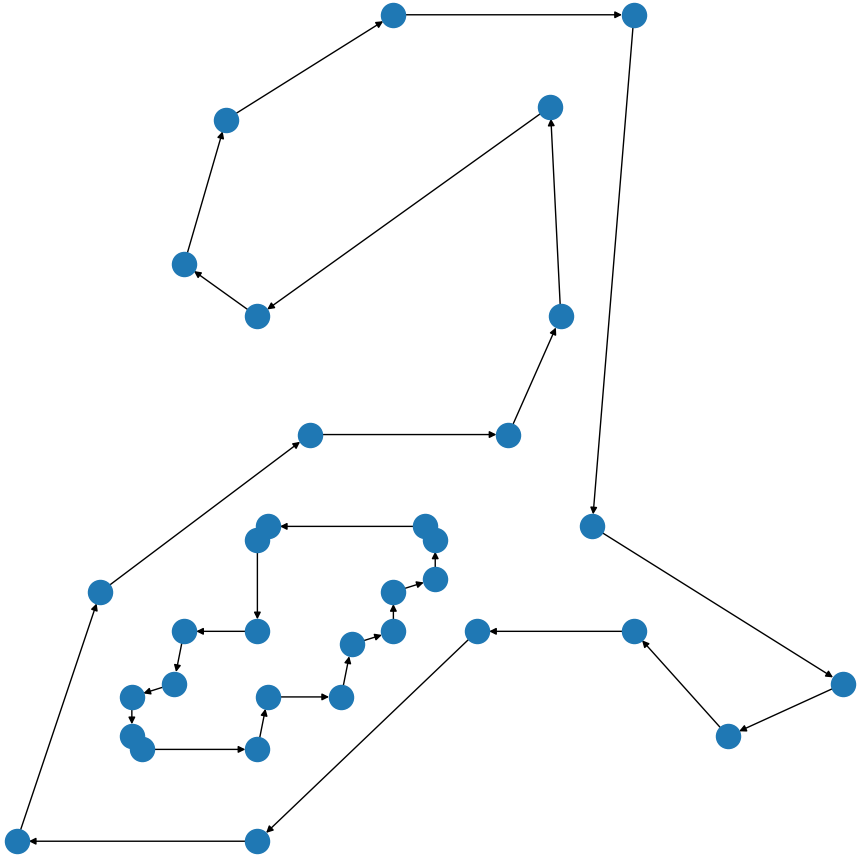}
         \caption{Approximated contours and corners.}
         \label{fig:representation_example_bcn}
     \end{subfigure}
     \hfill
     \begin{subfigure}[t]{0.25\textwidth}
         \centering
         \includegraphics[width=\textwidth]{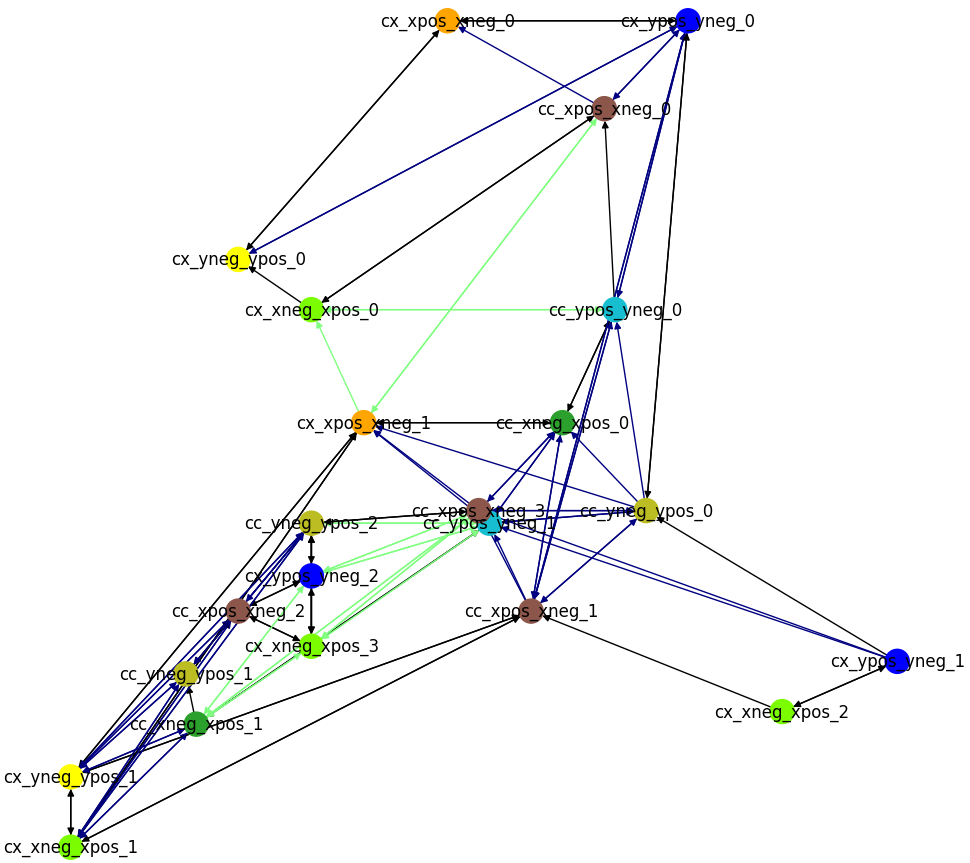}
         \caption{Final SPN, all networks drawn together.}
         \label{fig:representation_example_vn}
     \end{subfigure}
     \caption{Example SPN construction from an image for 2D shape identification. In (c), blue, green, and black edges denote inner, outer, and contour connections, respectively. Nodes, colored by type, represent gradient change points (e.g., "cc\_xneg\_xpos\_1" indicates a concave corner with x-gradient shifting from (-) to (+)).}
    \label{fig:feature_representation}
\end{figure}

To ensure generality in shape detection and avoid overly hand-crafted features, we use \textit{horizontal (x) and vertical (y) gradient orientation change points}. Our image processing flow involves: (1) converting a grayscale image to binary with a 50\% threshold, (2) approximating contours as polygons using OpenCV's Ramer–Douglas–Peucker algorithm \cite{opencv_shape_analysis}, yielding corners and oriented edges (CW for outer, CCW for inner contours) (Figure \ref{fig:representation_example_bcn}), and (3) computing the sign of gradients at each corner in the x and y dimensions based on edge orientation (e.g., a rightward-facing edge has a negative x-axis gradient).

We build the final SPN using corners where gradient orientations change in the x or y axes. Traversing the contour in its connected direction (CW or CCW), we create a node for each corner if the gradient direction (positive or negative) changes in either axis from the predecessor to the successor edge. The node type is defined by the change axis, direction, and corner convexity. For example, the top-right corner in Figure \ref{fig:representation_example_bcn} represents a convex corner with a y-gradient change from positive to negative, resulting in a node of type \textit{"convex, +y to -y"} (node \textit{cx\_ypos\_yneg\_0} in Fig. \ref{fig:representation_example_vn}). This process defines the SPN's nodes.

Our SPN includes four types of SNs: \textit{contour}, \textit{inner}, \textit{outer}, and \textit{all}. \textit{Contour} SNs represent connections along the contour, linking nodes as described earlier, with edges added between predecessors and successors of skipped corners. \textit{Inner} and \textit{outer} SNs connect nodes with straight lines that stay within the inner (pixel value 1) or outer (value 0) regions of the binary image. \textit{All}-type SNs combine all connections, regardless of type, to persistently represent relative positional relationships even when their actual types can vary across images. Each network type has \textit{horizontal} and \textit{vertical} variants, where directed edges indicate positional relationships along the respective axis. For example, a directed edge $(n_0, n_1)$ in \textit{contour-horizontal} means $n_0$ is to the left of $n_1$. See Fig. \ref{fig:representation_example_vn} for an example SPN.

\textit{Limitations and possible extensions:} This representation is broadly applicable and domain-agnostic for 2D shape detection, relying on contours matching image gradients. However, it’s mainly for demonstrating our design and has limitations, such as limited expressivity from not capturing all details of the shape but only those that represent gradient sign changes (e.g., only three of five corners are used in Fig. \ref{fig:representation_example_vn}). This affects final identification accuracy, making it fall short from a perfect performance as discussed in Section \ref{sec:results}. A more complete approach would consider all gradient orientation changes, increasing complexity but also completeness. Likewise, it currently works for shape detection, but could be extended to other 2D features or 3D spaces with similar logic. Additionally, for learning with a varsel mechanism, shape detection might benefit from alternative representations like pixel-level processing, CNN filters, or frequency-based transformations, as discussed in the Conclusion. Finally, we note that multiple feature types can be used with SPN representation, either in separate networks or within the same network, capturing positional relations between features across domains. While we don't explore this here, it’s an interesting direction for future work.

\subsection{Details of experimental framework}
\label{sec:appendix_expsetup}

\paragraph{Computation resources} All experiments were run on a 2.4GHz 8-Core Intel Core i9 processor with 32 GB 2667MHz DDR4 memory. No GPU was used. Giving an accurate estimate for computation time is impossible since experiments were run in parallel to unevenly-distributed independent workloads.

\subsubsection{Modelleyen}

\begin{figure}[t]{}
     \centering
     \includegraphics[width=0.5\textwidth]{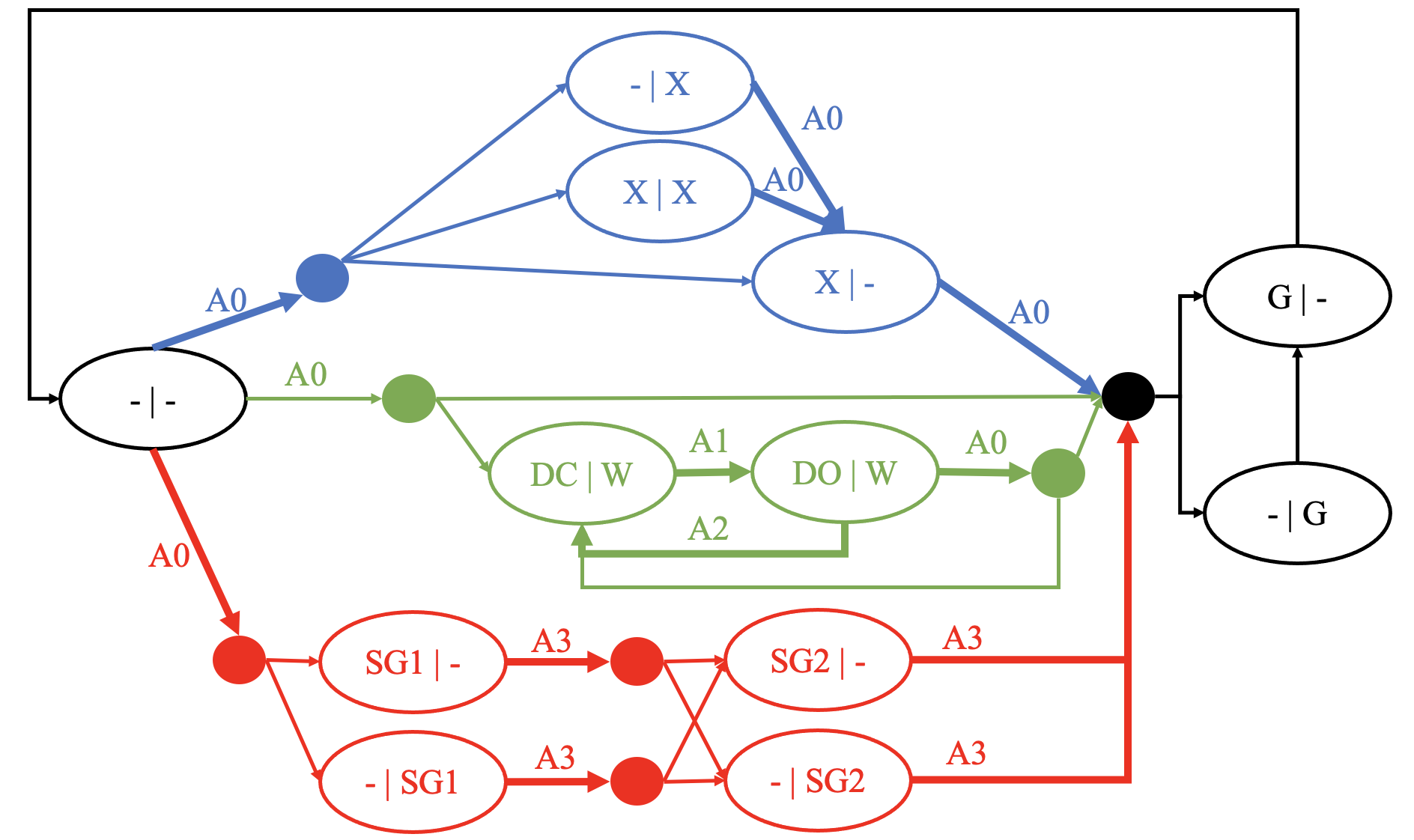}
     \caption{The environment and its subenvironments that we test on, essentially a FSM with two cells each of whom can take one of the states "DO, DC, W, G, SG1, SG2, X" or be empty ("-"). Each state is connected with arrows representing succession relations between them; filled circles correspond to multiple alternatives that can result from it. Green, red and blue portions are "RS", "SG", and "NEG" subtypes respectively (detailed in text), black portion is included in all subtypes. In "Complete" variant, all transitions and states are included. The agent’s goal is to activate state "G" in the first cell, and optimal actions are indicated by bold transitions. The environment has 20 actions, much larger than what is actually useful, in order to make it difficult to reach goal randomly.}
     \label{fig:environment}
 \end{figure}

\paragraph{Test environment} Figure \ref{fig:environment} shows the test environment used for our behavior experiments. The environment includes three subtypes ("RS", "SG", "NEG"), illustrated by different colors. This setup was designed to model various types of temporal successions, such as basic succession, correlated changes, alternative causes/outcomes, uncertain transitions, and negative conditons. There is also a \textit{random} variant of the environment where two additional states that get activated randomly are introduced, in order to test statistical significance filtering mechanisms. The environment was inspired from Multiroom environment in Minigrid. The states represent closed door (DC), open door (DO), wall (W), subgoal 1/2 (SG1/2), goal (G) and a random variable (X); "RS" stands for "rooms" and represents an agent going through multiple rooms opening doors in each, and "SGS" represents one in which agent reaches two subgoals and then reaches the goal afterwards, and "NEG" represents a case where goal appears conditioned on one positive and one negative conditon. In all, the goal can be moving. Alternative outcomes are present in all environment subtypes, since each of them allows for multiple outcomes following an empty ("-/-") state. Alternative predecessors are tested in "SGS" environment where SG2 can be preceded by SG1 in either of the two cells; and likewise in general the appearance of G can be preceded by any of the alternatives associated with different environment subtypes. The capability to represent positive and negative relations together is tested in subtype "NEG", in which G appears only if X is enabled in the first cell and not the second one.

\paragraph{Significance filtering} Modelleyen's mechanism of filtering based on statistical significance (i.e. NCE) is enabled only for the random variant of the environment. When enabled, we used a cutoff NCE of 0.25 for blocking upstream conditioner formations (i.e. no more upstream conditioners are formed if the CSV does not cause a >25\% in the probability of occurrence of its target).

\subsection{MNR}
\label{sec:exp_details}

\paragraph{Sample sizes:} We use $N_{sample}=20,\ 10,\ 5$ and test set size of $50,\ 20,\ 10$ samples per class for experiments with $N_C=3,\ 5,\ 10$ respectively. Reported results are averages of 10 runs for $N_C=3$ and 5 runs for $N_C=5$. Population size for generating assignments was chosen as $10$ for both learning and prediction.

\paragraph{MNR settings:} For MNR, we choose refinement threshold $T_{ref}=0.05$, significance threshold $T_{sign}=0.05$. $\epsilon$ for polygonal approximation is $0.01L$ where $L$ is the arc length of contour being approximated.

\paragraph{Neural network settings:} Our fully connected neural network architecture is of 2 hidden layers with 128 neurons each, while the CNN architecture has a pair of convolutional (32 filters with 3x3 kernels) and max-pooling (with pool size 2x2) layer, repeated twice sequentially, followed by a dense layer of 128 neurons. All NNs use ReLU activation in hidden layers and softmax in output. We use a maximum of 100 epochs and an early stopping patience of 10 epochs. All remaining settings are Keras defaults.

\paragraph{Prediction in MNR} To predict targets for a given observed SPN, we follow this procedure: First, we attempt to find an assignment that \textit{satisfies} the source SPN of the CSV. If no assignment is found, the CSV is considered inactive. If an assignment is found, we compute the activation probability of the CSV. If the CSV is unconditional (no positive/negative conditioners), the probability is $p = P(I(T)|SS(C))$, tracked over the learning process. If the CSV is conditional, the probability is $p = (1 - p^{max}-) \cdot p^{max}+$, where $p^{max}-$ and $p^{max}+$ are the maximum activation probabilities of negative and positive conditioners, respectively. Intuitively, this approach prioritizes the most-upstream representations (most closely matching the observed SPN) when calculating the final probability, resolving conflicts between activating and suppressing pathways by multiplying their probabilities.

\subsection{A sample model learned on SMR}

\begin{figure*}
  \centering
  \includegraphics[width=0.5\textwidth]{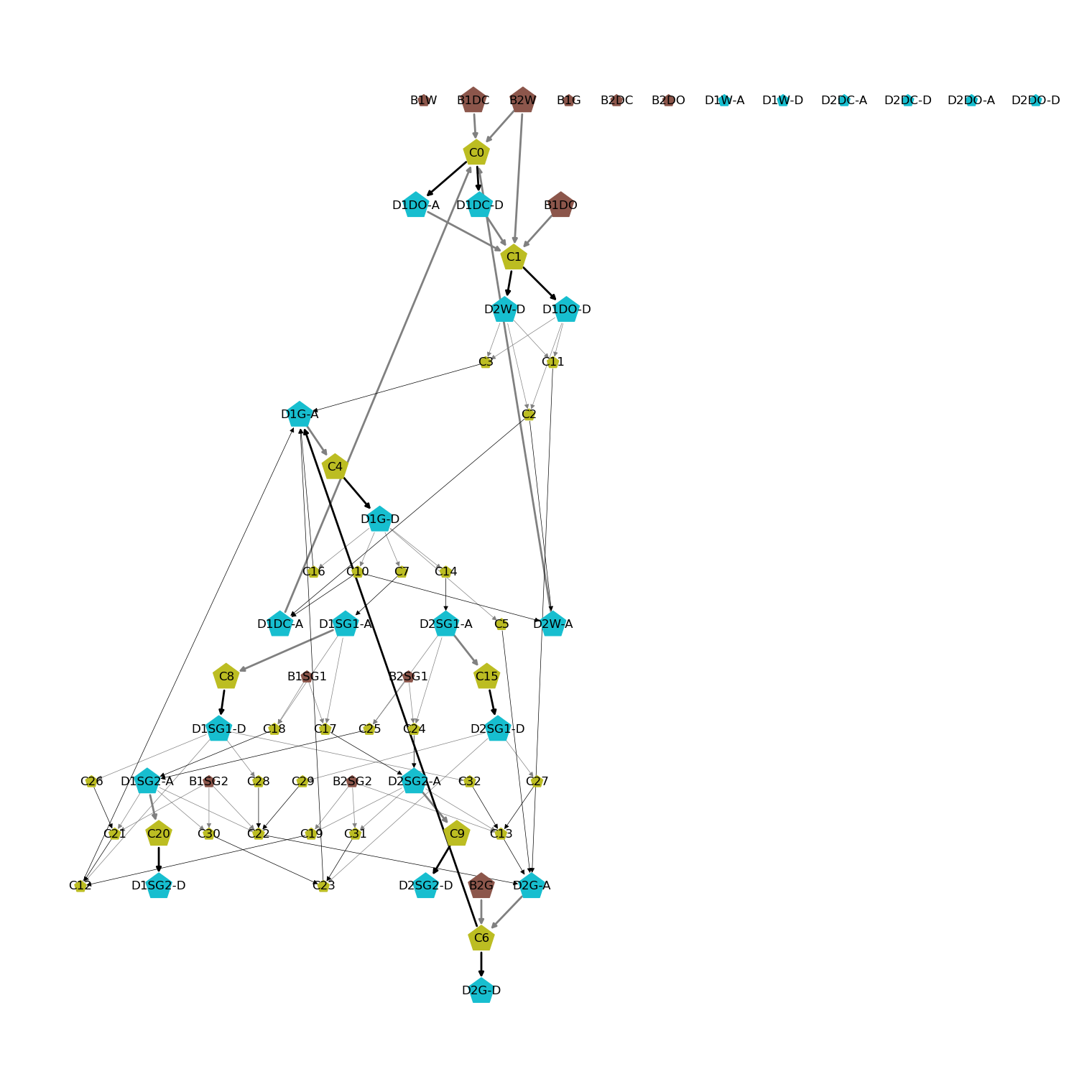}
  \caption{A sample environment model learned by Modelleyen. In the visualized model, brown nodes are BSVs, blues are DSVs, and the rest are CSVs. The enlarged pathways (bold arrows and large nodes) are reliable outcomes (i.e. unconditional CSVs) and the rest are uncertain (possibly conditional) ones. Black arrows represent conditioning relationships and gray arrows represent source relationships (all positive in this example). Disconnected SVs (those that can never be activated by environment design) are cut for visual clarity.}
  \label{fig:sample_model}
\end{figure*}

\begin{figure}
  \centering
  \includegraphics[width=0.5\textwidth]{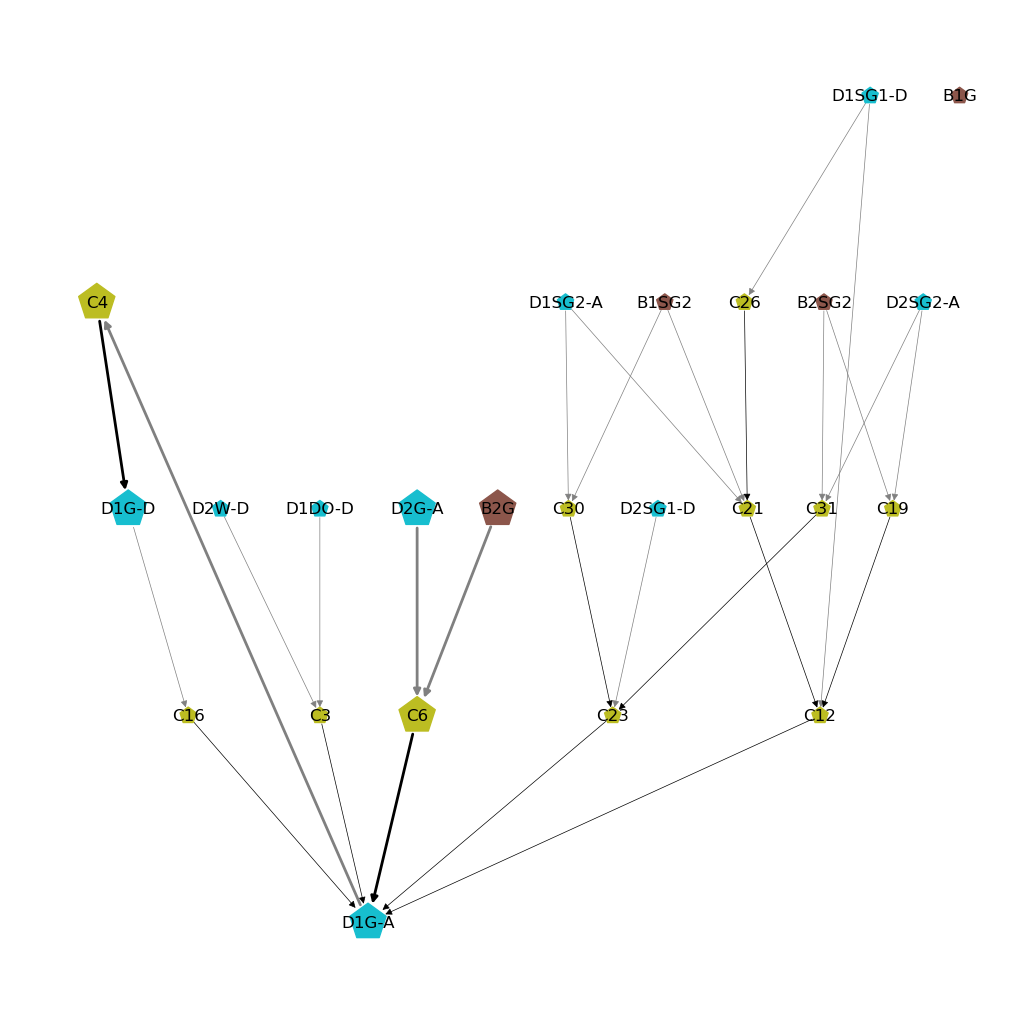}
  \caption{Same model as Figure \ref{fig:sample_model}, but for the predictive pathway of BSV 1G only. Many pathways for the activation of 1G can be seen in a human-comprehensible way in this model via the distinct CSVs preceding it (C3, C6, C12 C16, C23) and that the only reliable one of them is C6, and whose further sources can be seen by pursuing them upstream. In contrast, interpretation of a neural network model is much less straightforward due to nonlinearities, continuous parameters, and extensive connectivity that ties each neuron at the output to virtually all other neurons in the network.}
  \label{fig:sample_model_goal}
\end{figure}

\begin{figure*}
  \centering
  \includegraphics[width=0.5\textwidth]{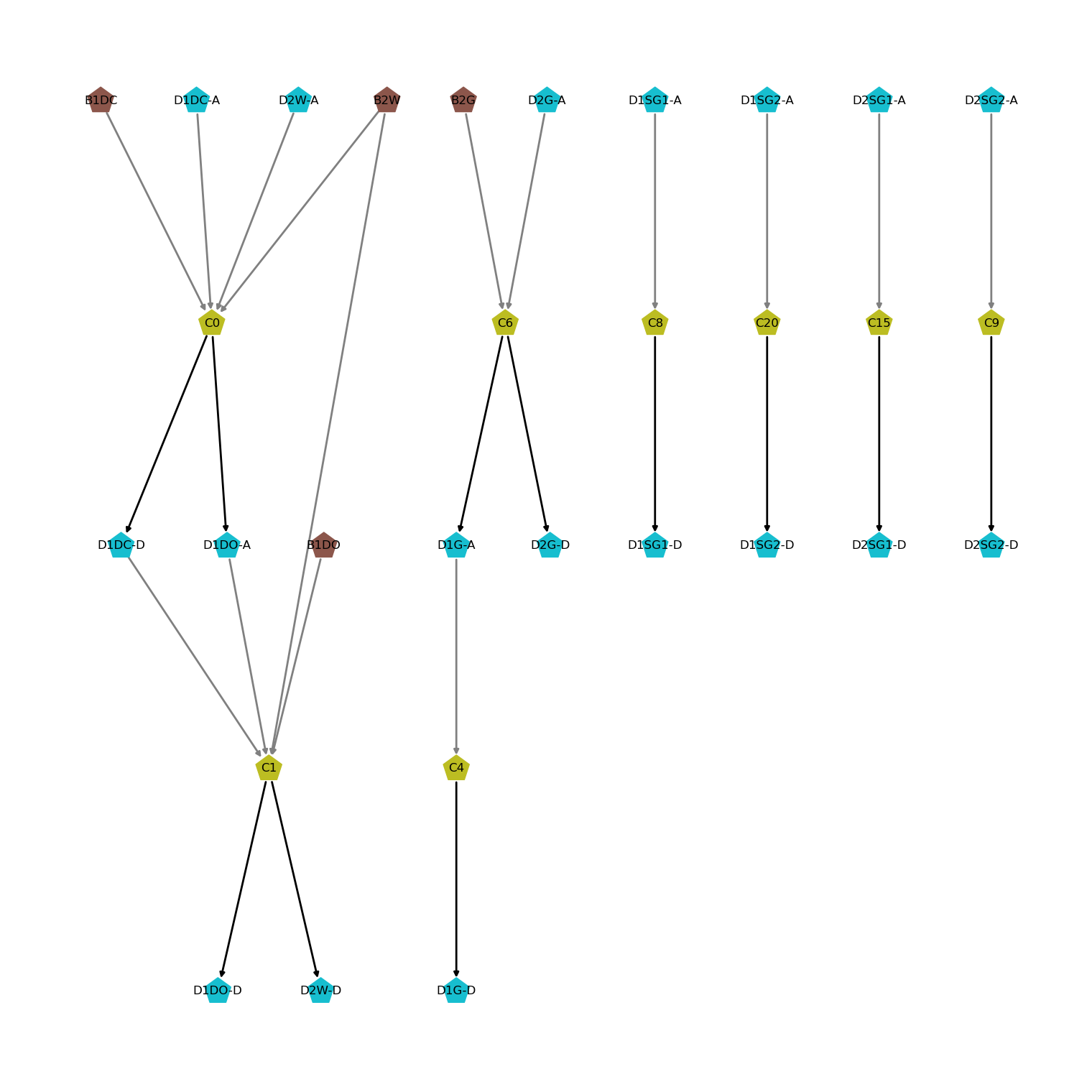}
  \caption{Same model as Figure \ref{fig:sample_model}, but with reliable pathways only, showing "islands of certainty" as potential candidates for abstraction.}
  \label{fig:sample_model_reliable}
\end{figure*}

A sample model learned on the SMR environment (Figure \ref{fig:environment}) is provided on Figure \ref{fig:sample_model}. Figure \ref{fig:sample_model_goal} provides, as an example, the pathway of BSV 1G (state G at cell 1), in which the specific pathways connecting to this BSV can be seen more clearly in a human-comprehensible manner. Figure \ref{fig:sample_model_reliable} shows the whole model, but only with reliable connections; clearly showing "islands of certain state transitions" which can be an example of a delimiting criterion that can be used for abstractions.

\begin{algorithm}[tb]
\caption{Simplified overview of the planning algorithm, relying on recursive generation of upstream \textit{action networks} (the graph of behaviors required to realize the desired goals from the currently active SVs).}

\label{alg:planner}
\textbf{Function} Plan(currentActiveSVs, goalSVs)
\begin{algorithmic}[1] 
    \STATE ActionNetwork $\leftarrow$ \ EmptyNet
    \FOR{SV, target $\in$ goalSVs}
        \STATE GenerateUpstreamAN(SV, target)
    \ENDFOR
\end{algorithmic}

\textit{Comment: Argument "target" states what the desired state is in the SV, which can be activation (A), deactivation (D), active (1) or nonactive (0). Irrelevant for CSVs.}

\textbf{Function} GenerateUpstreamAN(SV, target)
\begin{algorithmic}[1] 
    \STATE \textbf{if} satisfiedByCurrentActives(SV, target): return \textbf{True} 
    \STATE pathways $\leftarrow$ EmptyList
    \IF{type(SV) in [BSV, GSV]}
        \STATE pathways.add(Precondition(sv, target))
        \STATE{\textit{Comment: These are the preconditions for target to occur in a SV. For (A, D, 1, 0) they are (0, 1, A, D) respectively; since a SV must be activated for itself to be active, needs to be inactive for itself to get activated, and so on.}}
        \STATE pathways.add(Constituents(sv), target)
        \STATE pathways.add(Constituencies(sv), target)
        \STATE \textbf{if} target in ['A', 'D']: pathways.add(Conditioners(sv, target))
    \ELSIF{type(SV) is CSV}
        \STATE pathways.add(Sources(sv))
        \STATE pathways.add(Conditioners(sv))
    \ENDIF
    \STATE \textbf{if} pathways is Empty: return \textbf{False}

    \FOR{upstreamSV, upstreamTarget in pathways}
        \STATE ActionNetwork.AddEdge((upstreamSV, upstreamTarget), (SV, target))
        \STATE GenerateUpstreamAN(upstreamSV, upstreamTarget)
    \ENDFOR
\end{algorithmic}
\end{algorithm}

\subsection{Results on comprehensibility of learned representations by MNR}
\label{sec:mnr_results_comprehensibility}

\begin{figure*}
     \centering
     \begin{subfigure}{0.25\textwidth}
         \centering
         \includegraphics[width=\textwidth]{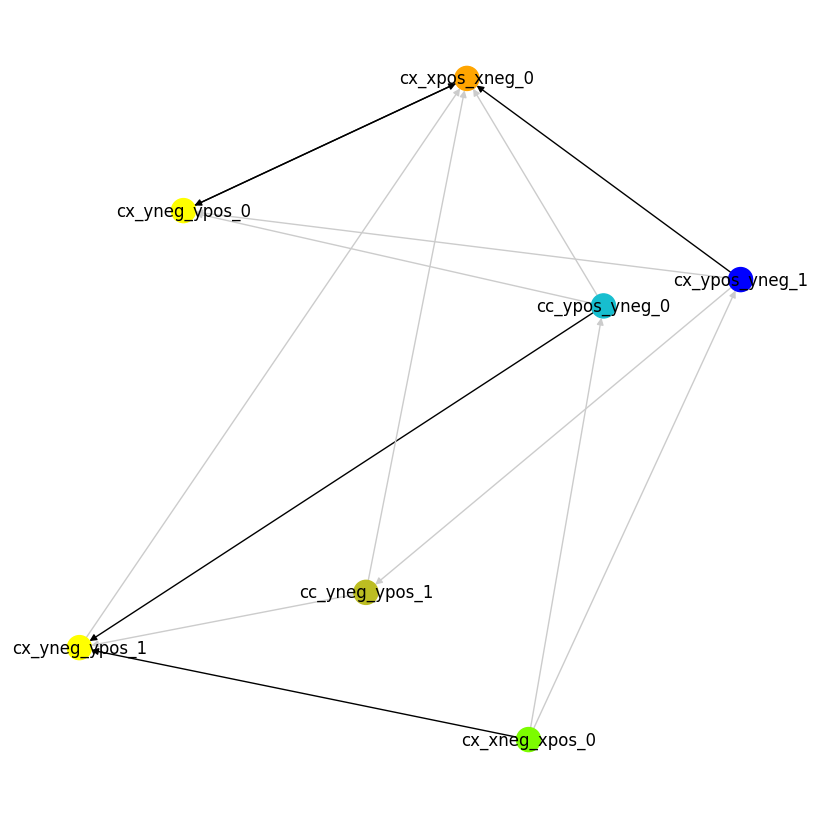}
         \caption{Digit 2, depth 0.}
         \label{fig:results_spns_2_ds}
     \end{subfigure}
      \begin{subfigure}{0.25\textwidth}
         \centering
         \includegraphics[width=\textwidth]{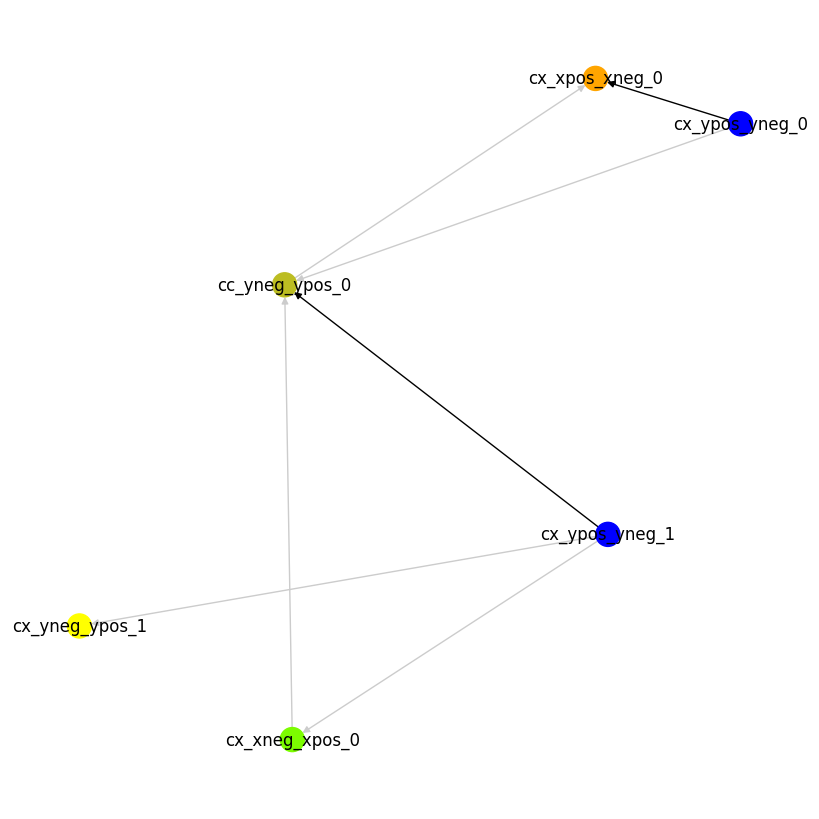}
         \caption{Digit 5, depth 0.}
         \label{fig:results_spns_5_ds}
     \end{subfigure}
     \begin{subfigure}{0.25\textwidth}
         \centering
         \includegraphics[width=\textwidth]{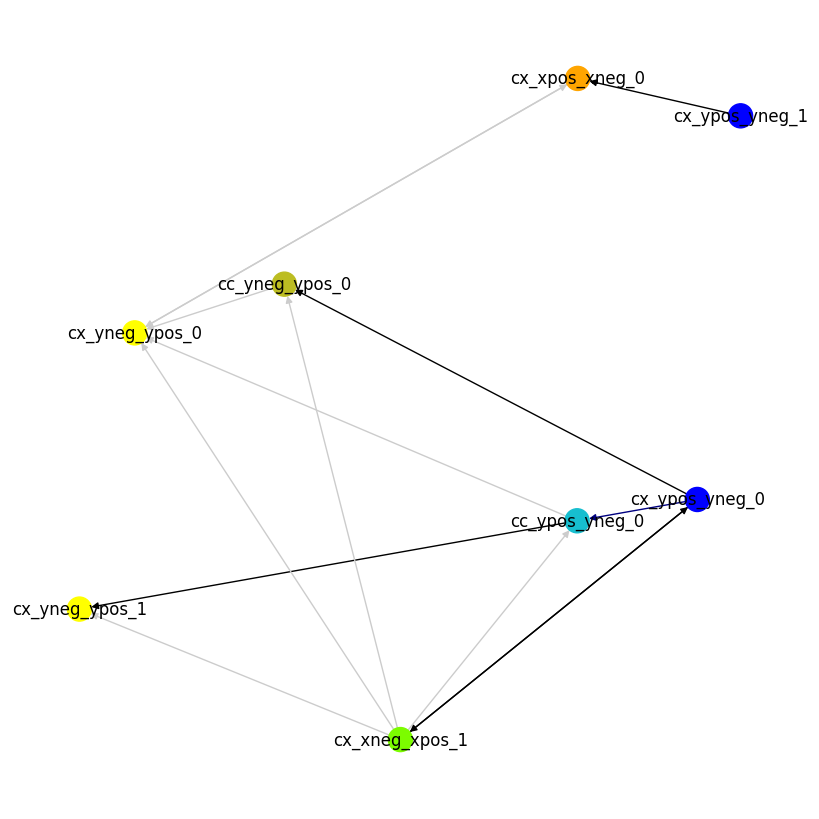}
         \caption{Digit 5, depth 8.}
         \label{fig:results_spns_5_us9}
     \end{subfigure}
    \caption{Sample source SPNs learned by the system, shown with all SNs combined. Blue, black, and gray edges represent inner, contour, and all-type connections, respectively. Depths indicate conditioning distance from the target variable (number of CSVs in the conditioning path). Node positions reflect the average of all observed positions during the CSV's lifetime. Additional examples are in Fig. \ref{fig:results_spns_appendix}.}
    \label{fig:results_spns}
\end{figure*}

\begin{figure*}
     \centering
     \begin{subfigure}{0.45\textwidth}
         \centering
         \includegraphics[width=\textwidth]{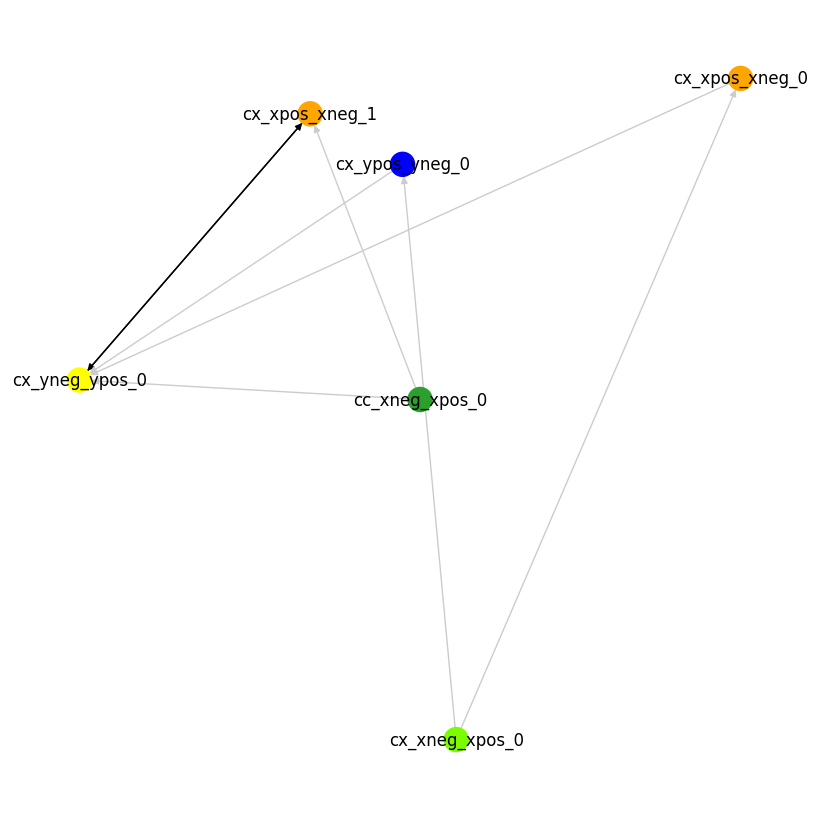}
         \caption{Digit 4, depth 0.}
         \label{fig:results_spns_4_ds}
     \end{subfigure}
     \begin{subfigure}{0.45\textwidth}
         \centering
         \includegraphics[width=\textwidth]{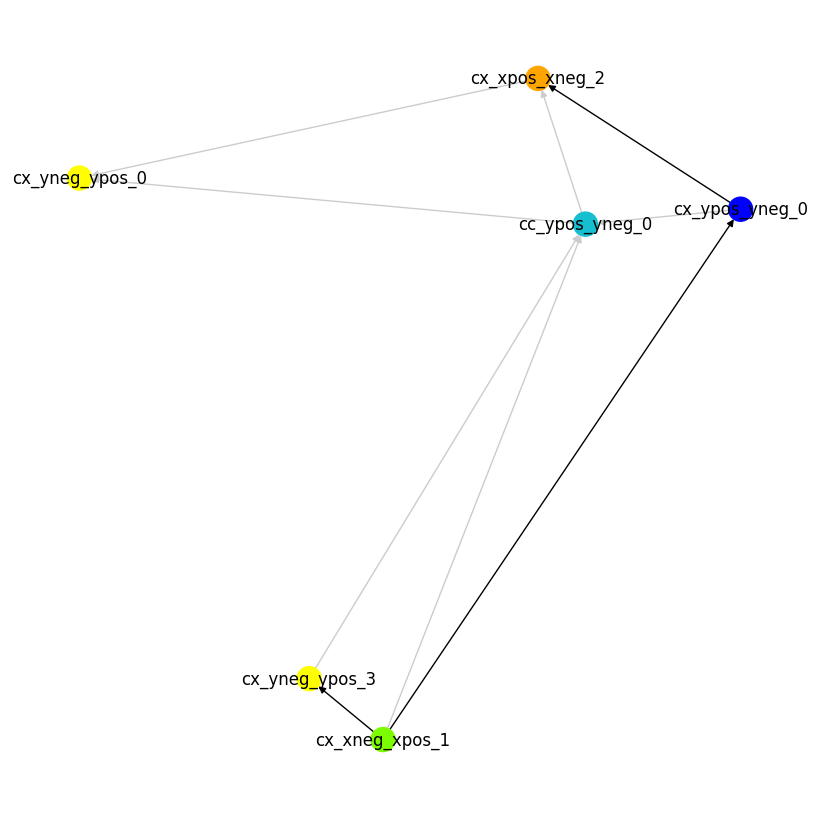}
         \caption{Digit 7, depth 0.}
         \label{fig:results_spns_7_ds}
     \end{subfigure}
      \begin{subfigure}{0.45\textwidth}
         \centering
         \includegraphics[width=\textwidth]{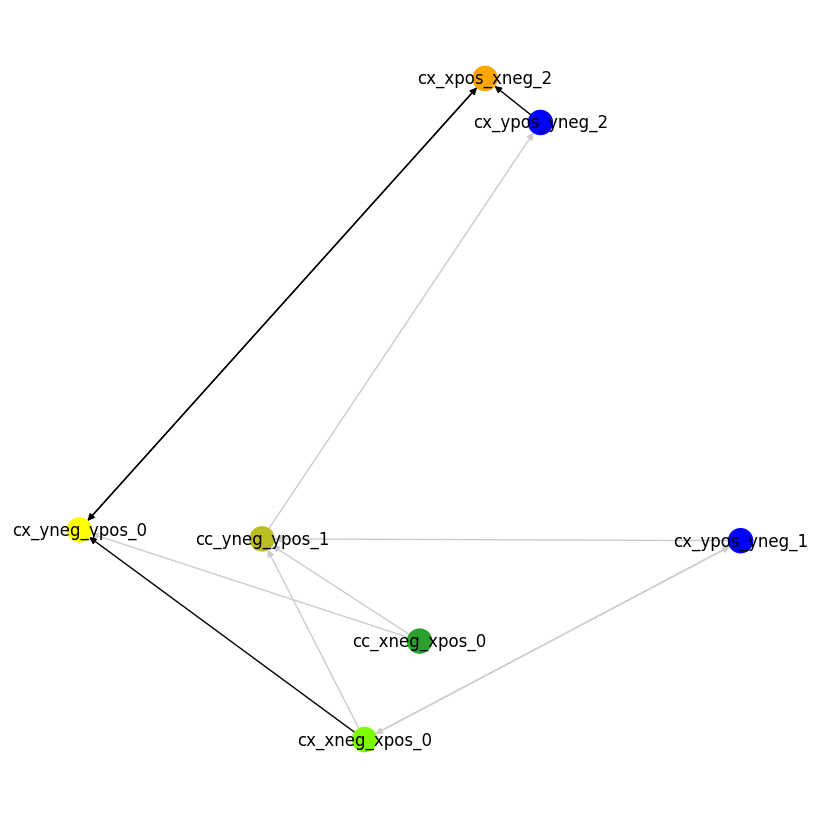}
         \caption{Digit 6, depth 0.}
         \label{fig:results_spns_6_ds}
     \end{subfigure}
     \begin{subfigure}{0.45\textwidth}
         \centering
         \includegraphics[width=\textwidth]{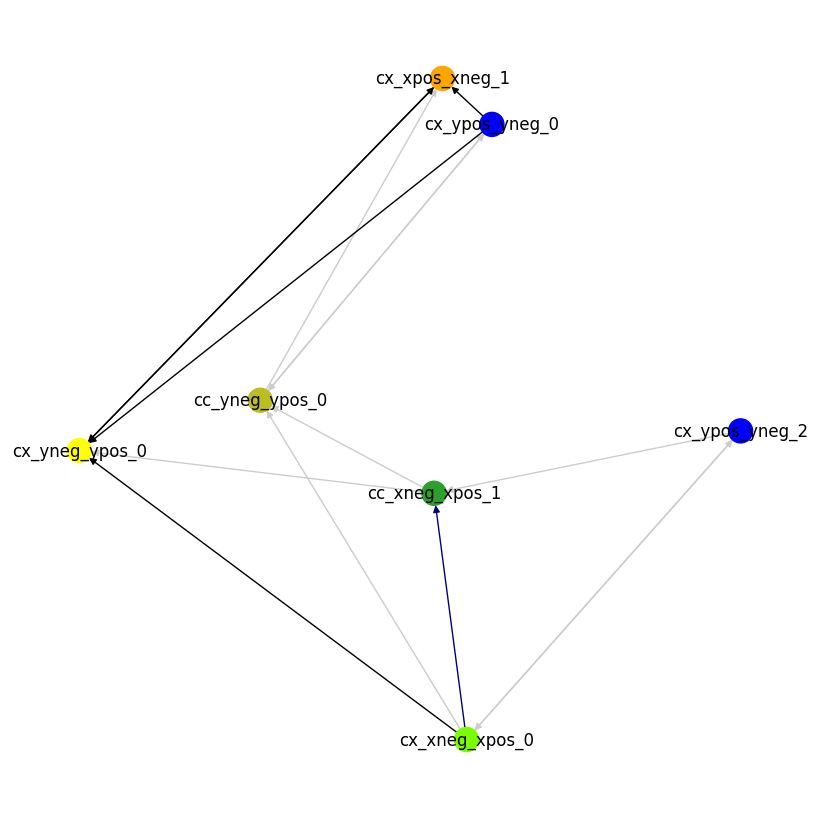}
         \caption{Source of a negative conditioner (suppressing connection) for digit 6, initialized from an instance of digit 4.}
         \label{fig:results_spns_6_negfrom4}
     \end{subfigure}
    \caption{Additional examples of representations of digits learned by MNR.}
    \label{fig:results_spns_appendix}
\end{figure*}

Figure \ref{fig:results_spns} illustrates samples of the learned SPNs, ranging from general representations at lower depths (near the target variable) to more specific ones at higher depths capturing rarer subvariants. These representations are visually intuitive, effectively depicting the digits, their features, and interrelations. For instance, most contours of digit "2" are preserved in Fig. \ref{fig:results_spns_2_ds}, though features like holes at the lower-left turning point (common in some samples like that in Fig. \ref{fig:feature_representation}) are omitted, while persistent features, such as the vertical gradient change (\textit{"cx\_yneg\_ypos\_1"}), are retained. Similarly, digit "5" in Fig. \ref{fig:results_spns_5_ds} retains key features, including vertical gradient changes on the right and horizontal changes at the top and bottom, along with correct positional relations. While general contours of "5" are refined at depth 0, they are preserved at the more specific subvariants upstream, like in Fig. \ref{fig:results_spns_5_us9} which provide more details. Some additional examples are also provided in Figure \ref{fig:results_spns_appendix}.

\begin{thebibliography}{27}
\providecommand{\natexlab}[1]{#1}
\providecommand{\url}[1]{\texttt{#1}}
\expandafter\ifx\csname urlstyle\endcsname\relax
  \providecommand{\doi}[1]{doi: #1}\else
  \providecommand{\doi}{doi: \begingroup \urlstyle{rm}\Url}\fi

\bibitem[Buzzega et~al.(2020)Buzzega, Boschini, Porrello, Abati, and Calderara]{buzzega2020dark}
Pietro Buzzega, Matteo Boschini, Angelo Porrello, Davide Abati, and Simone Calderara.
\newblock Dark experience for general continual learning: a strong, simple baseline.
\newblock \emph{Advances in neural information processing systems}, 33:\penalty0 15920--15930, 2020.

\bibitem[Clune(2019)]{clune2019ai}
Jeff Clune.
\newblock Ai-gas: Ai-generating algorithms, an alternate paradigm for producing general artificial intelligence.
\newblock \emph{arXiv preprint arXiv:1905.10985}, 2019.

\bibitem[Cortes(1995)]{cortes1995support}
Corinna Cortes.
\newblock Support-vector networks.
\newblock \emph{Machine Learning}, 1995.

\bibitem[Edelman(1993)]{edelman1993neural}
Gerald~M Edelman.
\newblock Neural darwinism: selection and reentrant signaling in higher brain function.
\newblock \emph{Neuron}, 10\penalty0 (2):\penalty0 115--125, 1993.

\bibitem[Erden \& Faltings(2024)Erden and Faltings]{erden2024directed}
Zeki~Doruk Erden and Boi Faltings.
\newblock Directed structural adaptation to overcome statistical conflicts and enable continual learning.
\newblock In \emph{AAAI 2024}, Deployable AI Workshop, 2024.

\bibitem[Galashov et~al.(2023)Galashov, Mitrovic, Tirumala, Teh, Nguyen, Chaudhry, and Pascanu]{galashov2023continually}
Alexandre Galashov, Jovana Mitrovic, Dhruva Tirumala, Yee~Whye Teh, Timothy Nguyen, Arslan Chaudhry, and Razvan Pascanu.
\newblock Continually learning representations at scale.
\newblock In \emph{Conference on Lifelong Learning Agents}, pp.\  534--547. PMLR, 2023.

\bibitem[Gerhart \& Kirschner(2007)Gerhart and Kirschner]{gerhart2007theory}
John Gerhart and Marc Kirschner.
\newblock The theory of facilitated variation.
\newblock \emph{Proceedings of the National Academy of Sciences}, 104\penalty0 (suppl\_1):\penalty0 8582--8589, 2007.

\bibitem[Ghallab et~al.(2016)Ghallab, Nau, and Traverso]{ghallab2016automated}
Malik Ghallab, Dana Nau, and Paolo Traverso.
\newblock \emph{Automated planning and acting}.
\newblock Cambridge University Press, 2016.

\bibitem[Hadsell et~al.(2020)Hadsell, Rao, Rusu, and Pascanu]{hadsell2020embracing}
Raia Hadsell, Dushyant Rao, Andrei~A Rusu, and Razvan Pascanu.
\newblock Embracing change: Continual learning in deep neural networks.
\newblock \emph{Trends in cognitive sciences}, 24\penalty0 (12):\penalty0 1028--1040, 2020.

\bibitem[Jacobson et~al.(2022)Jacobson, Wright, Jiang, Rodriguez-Rivera, and Xue]{jacobson2022task}
Maxwell~J Jacobson, Case~Q Wright, Nan Jiang, Gustavo Rodriguez-Rivera, and Yexiang Xue.
\newblock Task detection in continual learning via familiarity autoencoders.
\newblock In \emph{2022 IEEE International Conference on Systems, Man, and Cybernetics (SMC)}, pp.\  1--8. IEEE, 2022.

\bibitem[Kirkpatrick et~al.(2017)Kirkpatrick, Pascanu, Rabinowitz, Veness, Desjardins, Rusu, Milan, Quan, Ramalho, Grabska-Barwinska, et~al.]{kirkpatrick2017overcoming}
James Kirkpatrick, Razvan Pascanu, Neil Rabinowitz, Joel Veness, Guillaume Desjardins, Andrei~A Rusu, Kieran Milan, John Quan, Tiago Ramalho, Agnieszka Grabska-Barwinska, et~al.
\newblock Overcoming catastrophic forgetting in neural networks.
\newblock \emph{Proceedings of the national academy of sciences}, 114\penalty0 (13):\penalty0 3521--3526, 2017.

\bibitem[LeCun(2022)]{lecun2022path}
Yann LeCun.
\newblock A path towards autonomous machine intelligence version 0.9. 2, 2022-06-27.
\newblock \emph{Open Review}, 62\penalty0 (1), 2022.

\bibitem[LeCun et~al.(1998)LeCun, Bottou, Bengio, and Haffner]{lecun1998gradient}
Yann LeCun, L{\'e}on Bottou, Yoshua Bengio, and Patrick Haffner.
\newblock Gradient-based learning applied to document recognition.
\newblock \emph{Proceedings of the IEEE}, 86\penalty0 (11):\penalty0 2278--2324, 1998.

\bibitem[Lee et~al.(2020)Lee, Ha, Zhang, and Kim]{lee2020neural}
Soochan Lee, Junsoo Ha, Dongsu Zhang, and Gunhee Kim.
\newblock A neural dirichlet process mixture model for task-free continual learning.
\newblock \emph{arXiv preprint arXiv:2001.00689}, 2020.

\bibitem[Lindeberg(2012)]{lindeberg2012scale}
Tony Lindeberg.
\newblock Scale invariant feature transform.
\newblock 2012.

\bibitem[Marc(2005)]{marc2005plausibility}
Kirschner Marc.
\newblock \emph{The plausibility of life}.
\newblock Yale University Press, 2005.

\bibitem[Marcus(2018)]{marcus2018deep}
Gary Marcus.
\newblock Deep learning: A critical appraisal.
\newblock \emph{arXiv preprint arXiv:1801.00631}, 2018.

\bibitem[Mordoch et~al.(2023)Mordoch, Juba, and Stern]{mordoch2023learning}
Argaman Mordoch, Brendan Juba, and Roni Stern.
\newblock Learning safe numeric action models.
\newblock In \emph{Proceedings of the AAAI Conference on Artificial Intelligence}, volume~37, pp.\  12079--12086, 2023.

\bibitem[{OpenCV}(2025)]{opencv_shape_analysis}
{OpenCV}.
\newblock Shape analysis - opencv documentation, 2025.
\newblock URL \url{https://docs.opencv.org/4.x/d3/dc0/group__imgproc__shape.html#ga0012a5fdaea70b8a9970165d98722b4c}.
\newblock Accessed: 2025-01-29.

\bibitem[Oquab et~al.(2023)Oquab, Darcet, Moutakanni, Vo, Szafraniec, Khalidov, Fernandez, Haziza, Massa, El-Nouby, et~al.]{oquab2023dinov2}
Maxime Oquab, Timoth{\'e}e Darcet, Th{\'e}o Moutakanni, Huy Vo, Marc Szafraniec, Vasil Khalidov, Pierre Fernandez, Daniel Haziza, Francisco Massa, Alaaeldin El-Nouby, et~al.
\newblock Dinov2: Learning robust visual features without supervision.
\newblock \emph{arXiv preprint arXiv:2304.07193}, 2023.

\bibitem[Qu et~al.(2021)Qu, Rahmani, Xu, Williams, and Liu]{qu2021recent}
Haoxuan Qu, Hossein Rahmani, Li~Xu, Bryan Williams, and Jun Liu.
\newblock Recent advances of continual learning in computer vision: An overview.
\newblock \emph{arXiv preprint arXiv:2109.11369}, 2021.

\bibitem[Rusu et~al.(2016)Rusu, Rabinowitz, Desjardins, Soyer, Kirkpatrick, Kavukcuoglu, Pascanu, and Hadsell]{rusu2016progressive}
Andrei~A Rusu, Neil~C Rabinowitz, Guillaume Desjardins, Hubert Soyer, James Kirkpatrick, Koray Kavukcuoglu, Razvan Pascanu, and Raia Hadsell.
\newblock Progressive neural networks.
\newblock \emph{arXiv preprint arXiv:1606.04671}, 2016.

\bibitem[Wang et~al.(2022)Wang, Liu, Duan, Kong, and Tao]{wang2022continual}
Zhen Wang, Liu Liu, Yiqun Duan, Yajing Kong, and Dacheng Tao.
\newblock Continual learning with lifelong vision transformer.
\newblock In \emph{Proceedings of the IEEE/CVF Conference on Computer Vision and Pattern Recognition}, pp.\  171--181, 2022.

\bibitem[West-Eberhard(2003)]{west2003developmental}
Mary~Jane West-Eberhard.
\newblock \emph{Developmental plasticity and evolution}.
\newblock Oxford University Press, 2003.

\bibitem[Xu et~al.(2020)Xu, Qin, Sun, Wang, Chen, and Ren]{xu2020learning}
Kai Xu, Minghai Qin, Fei Sun, Yuhao Wang, Yen-Kuang Chen, and Fengbo Ren.
\newblock Learning in the frequency domain.
\newblock In \emph{Proceedings of the IEEE/CVF conference on computer vision and pattern recognition}, pp.\  1740--1749, 2020.

\bibitem[Zador(2019)]{zador2019critique}
Anthony~M Zador.
\newblock A critique of pure learning and what artificial neural networks can learn from animal brains.
\newblock \emph{Nature communications}, 10\penalty0 (1):\penalty0 3770, 2019.

\bibitem[Zhuang et~al.(2020)Zhuang, Qi, Duan, Xi, Zhu, Zhu, Xiong, and He]{zhuang2020comprehensive}
Fuzhen Zhuang, Zhiyuan Qi, Keyu Duan, Dongbo Xi, Yongchun Zhu, Hengshu Zhu, Hui Xiong, and Qing He.
\newblock A comprehensive survey on transfer learning.
\newblock \emph{Proceedings of the IEEE}, 109\penalty0 (1):\penalty0 43--76, 2020.

\end{thebibliography}
\end{document}